\documentclass[review]{elsarticle}
\usepackage{hyperref}

\usepackage{graphics}
\usepackage{graphicx}
\newsavebox\CBox
\def\textBF#1{\sbox\CBox{#1}\resizebox{\wd\CBox}{\ht\CBox}{\textbf{#1}}}
\usepackage{amsmath,amssymb,amsfonts, commath}
\usepackage{tabularx}
\usepackage{subcaption} 
\usepackage{booktabs}
\usepackage{siunitx}
\usepackage[capitalise]{cleveref}
\crefformat{appendix}{#2#1#3}
\usepackage[symbol]{footmisc}
\usepackage{diagbox}
\usepackage{xspace}
\usepackage{longtable}
\usepackage{float}

\DeclareMathOperator*{\E}{\mathbb{E}}
\newcommand{\minus}{\scalebox{0.75}[1.0]{$-$}}
\newcommand{\exLL}{$\E_{\theta}\mathrm{(NLL)}$\xspace}
\newcommand{\varLL}{$\mathrm{Var_{\theta}(NLL)}$\xspace}

\newcommand{\ExNLLShort}{$\E_{\theta}\mathrm{(NLL)}$\xspace}
\newcommand{\VarNLLShort}{$\mathrm{Var_{\theta}(NLL)}$\xspace}
\newcommand{\ExNLLFull}{$\E{}_{\theta}{[\minus{\log{p(\textbf{x}_n^*|\theta)}}]}$\xspace}
\newcommand{\VarNLLFull}{$\text{Var}_\theta[\minus{\log{p(\textbf{x}_n^*|\theta)}}]$\xspace}
\newcommand{\nonDriftSS}{$S^{\bot{shift}}$\xspace}
\newcommand{\DriftSS}{$S^{shift}$\xspace}

\newcommand{\ExNLLBoldShort}{$\boldsymbol{\E_{\theta}\mathrm{(NLL)}}$\xspace}
\newcommand{\VarNLLBoldShort}{$\boldsymbol{\mathrm{Var_{\theta}(NLL)}}$\xspace}

\journal{Journal of Applied Soft Computing}

\begin{document}

\begin{frontmatter}

\title{Coalitional Bayesian Autoencoders: Towards explainable unsupervised deep learning}

\author{Bang Xiang Yong, Alexandra Brintrup}
\address{Institute for Manufacturing, University of Cambridge, UK}




\begin{abstract}
This paper aims to improve the explainability of Autoencoder's (AE) predictions by proposing two explanation methods based on the mean and epistemic uncertainty of log-likelihood estimate, which naturally arise from the probabilistic formulation of the AE called Bayesian Autoencoders (BAE). To quantitatively evaluate the performance of explanation methods, we test them in sensor network applications, and propose three metrics based on covariate shift of sensors: (1) G-mean of Spearman drift coefficients, (2) G-mean of sensitivity-specificity of explanation ranking and (3) sensor explanation quality index ($SEQI$) which combines the two aforementioned metrics. Surprisingly, we find that explanations of BAE's predictions suffer from high correlation resulting in \textit{misleading} explanations. To alleviate this, a "Coalitional BAE" is proposed, which is inspired by agent-based system theory. Our comprehensive experiments on publicly available condition monitoring datasets demonstrate the improved quality of explanations using the Coalitional BAE.

\end{abstract}

\begin{keyword}
\texttt{Explainable deep learning, Bayesian Autoencoders, condition monitoring, Industry 4.0, Internet of Things}
\end{keyword}

\end{frontmatter}


\section{Introduction}

\begin{figure}[htbp]
\centerline{\includegraphics[scale=.55]{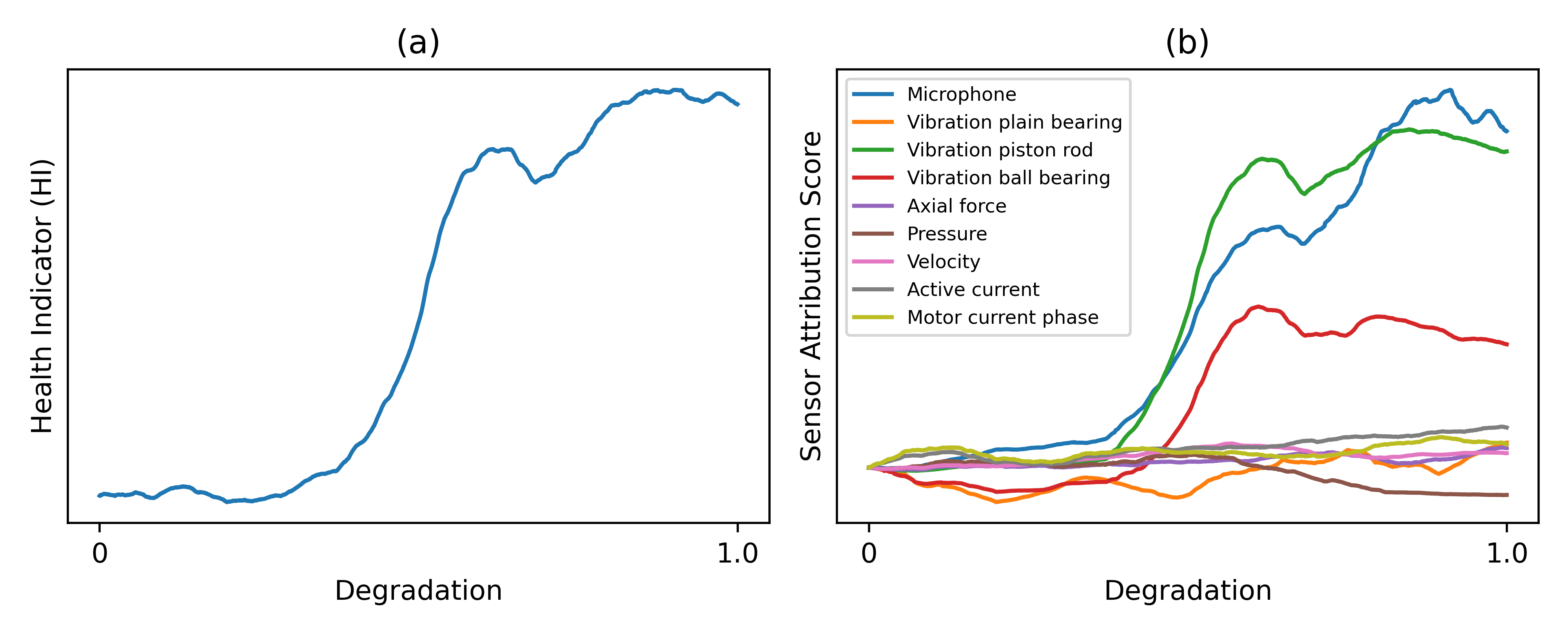}}
\caption{Illustrative examples of (a) health indicator of an equipment as the output of a DNN and (b) the local attribution scores for each sensor. At any degradation state (0 = healthy, 1 = total failure), the higher the sensor attribution score, the higher the importance of the sensor is to the model making the prediction.} 
\label{fig:example-sensor-attr}
\end{figure}

With recent advances in artificial intelligence and machine learning for Industrial Internet of Things (IIoT) devices, deep neural network (DNN) offers a promising way to analyse large amount of data for a range of applications. However, due to its black-box nature, DNN does not offer explainability out-of-the-box; for instance, given the model's prediction of the equipment's health in \cref{fig:example-sensor-attr}(a), one is unable to tell which sensors are contributing to the prediction without additional modelling effort to obtain explanations shown in \cref{fig:example-sensor-attr}(b). Being explainable is crucial to improving the trust and safety in DNN especially for a high stake industrial applications. 

One way of obtaining explanations for DNN predictions is by using posthoc explanation methods which use a separate model to assign attribution scores to the features. Nonetheless, a critical question remains - how good are these explanations? Recent studies in supervised DNN have shown the fragility of such explanations \cite{ghorbani2019interpretation}; despite the DNN achieving a high predictive accuracy, the explanations to these predictions may not be reliable. 

Extant explanation methods are mostly studied within the context of supervised DNN, and their success on unsupervised DNN is even more unclear. It is important to consider explainability for unsupervised DNNs too, as they have many applications in the field of manufacturing, in particular when considering unlabelled sensor data \cite{WANG2018144}. Without careful evaluation of the explanation methods for unsupervised DNN, manufacturers are exposed to the risk of \textit{misleading} explanations when adopting these methods. 

Therefore, in this paper, we focus our analysis on a manufacturing use-case: that of "sensor attribution scores" as a form of explanation obtained using Autoencoder (AE) which is a type of unsupervised DNN. Moreover, we adopt the probabilistic formulation of the AE called Bayesian Autoencoder (BAE) \cite{yong2020bayesian, yong2020icml} which provides a sound theoretical foundation for analysis. As a specific use-case, we study their applications in industrial equipment health indicator (HI). In particular, our contributions are as follows,

\begin{enumerate}
    \item \textbf{Formulation of two sensor attribution methods for BAE.} Under the independent likelihood assumption, we propose two methods for assigning attribution scores to sensor inputs which are based on (1) the mean log-likelihood estimate and (2) the epistemic uncertainty of log-likelihood estimate. We argue that the BAE predictions are \textit{naturally} decomposable into the attribution of features, without necessitating any posthoc explanation methods.
    \item \textbf{Development of "Coalitional BAE".} We propose the Coalitional BAE which is inspired by coalitional agent-based system theory to improve the explanation quality. In this setup, an independent BAE is trained for every sensor, as opposed to the conventional training of a Centralised BAE which combines all sensor inputs under a single BAE. 
    \item \textbf{Quantitative evaluation metrics for explainable HI based on covariate shift of sensors.} Considering that a HI is attributable to the sensors which are either shifting or non-shifting from the training distribution, we propose three metrics to evaluate how well do the explanations reflect the ground truth: (1) G-mean of Spearman drift coefficients ($G_{SDC}$), (2) G-mean of specificity-sensitivity of explanation ranking ($G_{SSER}$) and (3) sensor explanation quality index ($SEQI$) which is a weighted sum of the two aforementioned metrics. 
    \item \textbf{Finding of misleading explanations due to correlation in outputs.} Through our experiments, we find that the quality of explanation obtained with the Centralised BAE is poor due to correlation in the outputs. This sheds light on the propensity of Centralised BAE to give misleading explanations which was not noticed in extant studies. In contrast, by enforcing independence in the model outputs, the quality of explanation attained by Coalitional BAE consistently outperforms that of Centralised BAE.

\end{enumerate}


\section{Related work} \label{section-related-work}

\subsection{Health Indicator with Autoencoders}

In supervised DNN, the model learns a direct mapping from the inputs to the target labels, whereas in unsupervised learning, the model does not have access to the target labels and the model is trained only on the input data to infer the possible target labels. AE is a popular unsupervised DNN, and there exists two main schemes of obtaining a HI in the literature,

\begin{enumerate}
    \item \textbf{AE as an automatic feature extractor.} In this mode, the AE is trained to reconstruct the input data and subsequently a supervised model or clustering algorithm is trained to map the encoded features of lower dimensions to the target labels \cite{ren2018bearing}. Various modifications were proposed to improve the encoded features, such as stacked sparse AEs \cite{sohaib2018reliable, ma2018predicting}, enhanced regularisation method \cite{meng2018enhancement} and ensemble of stacked AEs \cite{lin2019novel}. Xu et al. \cite{XU2018898} improved the AE classification performance by applying Gath-Geva clustering algorithm on the encoded features.
    \item \textbf{Anomaly detection.} In condition monitoring applications, AEs are trained only to reconstruct the sensor measurements during the equipment's healthy phase. During test phase, measurements from a degrading equipment are assumed to deviate from the healthy training distribution and as such, would be assigned a higher reconstruction loss. In separate studies, the reconstruction loss has been used to differentiate healthy and damaged states of a bridge \cite{anaissi2019multi} and surface-mounted devices \cite{oh2018residual}. Amarbayasgalan et al. \cite{amarbayasgalan2020unsupervised} also studied the use of reconstruction loss in time-series anomaly detection. Recently, Yong et al. \cite{yong2020bayesian} proposed a new variant of AE called Bayesian Autoencoder (BAE), and studied the use of uncertainty and reconstruction loss to detect various types of drifts in a hydraulic system. 
\end{enumerate}

In this work, we investigate the explainability of HI obtained with the BAE trained under the anomaly detection mode and build upon the work of Yong et al. \cite{yong2020bayesian} on using BAE. 

\subsection{Explainable Deep Learning}

A recent review by Zhang et al. \cite{zhang2020deep} highlights the importance of explainability of DNN as a  future direction for its application in industry. To explain predictions of DNN, extant methods assign importance scores to the input features. Ribeiro et al. \cite{ribeiro2016should} proposed a model-agnostic method called Local Interpretable Model-agnostic Explanations (LIME) which uses a model of lower complexity to compute interpretable approximation of the original model. There are also DNN-specific explanation models such as DeepLift \cite{shrikumar2017learning} and IntegratedGradients \cite{sundararajan2017axiomatic} which leverages backpropagation to assign importance scores to the features based on the activation of the neurons. 

Although these methods were designed for supervised learning, Antwarg et al. \cite{antwarg2019explaining} studied the use of Shapley Additive Explanation (SHAP) \cite{lundberg2017unified} and LIME, to interpret the contribution of input features to the AE's reconstruction loss. In contrast, Bergmann et al. \cite{bergmann2019mvtec} proposed to directly use the reconstruction loss pertaining every input feature as the feature attribution score. Martinez-Garcia et al. \cite{martinez2019visually} proposed an entropy-based method to provide visual interpretations of the AE's reconstructed signals in turbine health monitoring. Liznerski et al. \cite{liznerski2020explainable} developed Fully Convolutional Data Description (FCDD) as a method to provide explanations to deep one-class classifier in a semi-supervised setting. 

One challenge in explainable deep learning is in evaluating the explanation methods and there has been a lack of consensus in extant studies since explainability is a subjective concept. In assessing these methods, qualitative evaluation is usually adopted by visualising the saliency map \cite{martinez2019visually, liznerski2020explainable}. However, it is also important to consider quantitative methods which enable more rigorous evaluation and to improve the explanation methods objectively. For this reason, Yeh et al. \cite{yeh2019fidelity} proposed the notion of fidelity and sensitivity as performance metrics of explanation methods. In addition, Hooker et al. \cite{hooker2019benchmark} presented a framework for Benchmarking Attribution Methods (BAM) with a priori knowledge of relative feature importance in the dataset which serves as ground truth. To the best of our knowledge, however, there has not been any comprehensive studies which use quantitative metrics to evaluate the explanation methods for unsupervised deep learning.

\section{Methods} \label{section-methods}
\subsection{Bayesian Autoencoder (BAE)}

This section describes the notations necessary to formulate the AE from a Bayesian perspective and the methods for training and predictions.  

\subsubsection{Notation}
Assume that on a piece of industrial equipment, we have a set of $K$ sensors, $S_{K}=\{{s}_k\}_{k=1}^K$. Let training data $X = \{\textbf{x}_n\}_{n=1}^{N}$ , $\textbf{x}_n \in \rm I\!R^{K\times{D}}$ where $D$ is the number of raw measurements or preprocessed features for the $K$ sensors. We further denote $x_{k,d}$ as an entry of data point $\textbf{x}_n$ for the $k$-th sensor and the $d$-th feature. 

The AE is a neural network parameterised by $\theta$ which maps $X$ into a set of reconstructed signals with the same dimensions as the input data point, $\hat{X}=\{\hat{\textbf{x}}_n\}_{n=1}^{N}$, $\hat{\textbf{x}_n} \in \rm I\!R^{K\times{D}}$. The AE consists of two parts: an encoder $f_\text{encoder}$, for mapping input data $\textbf{x}$ to the latent features of lower dimensional space, and a decoder $f_\text{decoder}$ for mapping the latent features to a reconstructed signal of the input $\hat{\textbf{x}}$ (i.e. $\hat{\textbf{x}} = {f_{\theta}}(\textbf{x}) = f_\text{decoder}(f_\text{encoder}(\textbf{x}))$) \cite{goodfellow2016deep}. 

\subsubsection{Training the BAE}

Bayes' rule can be applied to the parameters of the AE, to create a BAE,
\begin{equation}\label{eq_posterior}
    p(\theta|X) = \frac{p(X|\theta)\  p(\theta)}{p(X)} \\ ,
\end{equation}
where $p(X|\theta)$ is the likelihood and $p(\theta)$ is the prior distribution of the AE parameters. Following Yong et al. \cite{yong2020icml}, we use the isotropic Gaussian distribution with fixed variance of 1 to have independent likelihoods and a non-informative prior.

Since \cref{eq_posterior} is analytically intractable for a DNN which is highly non-linear and consists of a large number of parameters, we employ the state-of-the-art randomised maximum a posteriori (MAP) sampling method, which was developed recently by Pearce et al. \cite{pearce2020uncertainty} called \textit{anchored ensembling} for obtaining approximate samples from the posterior distribution. One major advantage of this method is the lower number of posterior samples ($\geq{5}$), as compared to Markov Chain Monte Carlo (MCMC) and mean field variational inference \cite{blundell2015weight} which typically require thousands of samples. In addition, it was shown empirically that ensembling DNNs typically yields good quality of predictive uncertainty \cite{lakshminarayanan2017simple, yao2019quality} and the \textit{anchored ensembling} grounds this technique in Bayesian theory \cite{pearce2020uncertainty}.

During initialisation, we independently sample a set of anchored weights $\{\theta_{anc,m}\}_{m=1}^{M}$ for the ensemble of $M$ AEs from the Kaiming uniform distribution \cite{he2015delving} and these anchored weights remain fixed throughout the training procedure. For each $m$-th AE in the ensemble, we minimise the following loss function using optimisers for DNN such as Adam \cite{kingma2014adam}, 

\begin{equation} \label{eqn:bae-loss}
Loss_{m} = \frac{1}{N}\sum^{N}_{n=1}{||\textbf{x}_n-f_{\theta_m}(\textbf{x}_n)||^2} + \frac{\lambda}{N}{||\theta_m-\theta_{anc,m}||^2}
\end{equation}
where $\lambda$ is set as a hyperparameter to scale the prior term for regularisation. Importantly, we note that minimising the mean-squared error loss which is often used as the AE reconstruction loss, is equivalent to maximising the log-likelihood of a isotropic Gaussian distribution with a fixed variance of 1. 
As a result of the training procedure, we obtain a set of approximate samples from the posterior distribution $\{{\hat{\theta{}_m}}\}_{m=1}^{M}$. 

\subsubsection{Predicting with BAE} \label{method-prediction-bae}

During the test phase, we use the ensemble to obtain $m$-samples of the negative log-likelihood for a test data point $\{\minus{\log{p({\textbf{x}_n^*}|\hat{\theta_m)}}\}}_{m=1}^{M}$ and subsequently compute the mean and epistemic uncertainty of the negative log-likelihood estimate with respect to the posterior distribution, 

\begin{equation}
\E{}_{\theta}{[\minus{\log{p(\textbf{x}_n^*|\theta)}}]} \approx{} \frac{1}{M}\sum_{m=1}^{M}{\minus{\log{p({\textbf{x}_n^*}|\hat{\theta_m)}}}}
\end{equation}
\begin{equation}
\text{Var}_\theta[\minus{\log{p(\textbf{x}_n^*|\theta)}}] \approx{} \frac{1}{M}\sum_{m=1}^{M}{(\minus{\log{p({\textbf{x}_n^*}|\hat{\theta_m)}}} -\E{}_{\theta}{[\minus{\log{p(\textbf{x}_n^*|\theta)}}]}})^2
\end{equation}
Both of these measures can be used for measuring dataset shifts. For brevity, we shall refer to \ExNLLFull and \VarNLLFull as \ExNLLShort and \VarNLLShort respectively. Intuitively, the higher the \ExNLLShort or \VarNLLShort, the greater the shift from the training distribution and the more severe the outcome \cite{yong2020bayesian}.

\subsubsection{Sensor Attribution Scores}
\label{section:sensor-attr}

The task of explaining a BAE prediction can be viewed as assigning attribution scores to $K$ sensors, given the test data point $\textbf{x}_n^*$. We denote the explanation method
$\Phi{(\textbf{x}_n^* {;}\: \theta{})} : \rm I\!R^{K\times{D}} \to \rm I\!R^{K} $ and the $k$-th sensor attribution score $\phi_{n,k} \in \Phi{(\textbf{x}_n^* {;}\: \theta{})}$.


We postulate that the BAE is naturally equipped with explanation methods to obtain these attribution scores and does not necessitate the use of posthoc explanation methods. Under the independent likelihood assumption, the overall log-likelihood is the sum of the log-likelihoods due to $D$ features and $K$ sensors,
\begin{equation}
\log{p(\textbf{x}^*_n|\theta{})} = \sum_{k=1}^K\sum_{d=1}^{D}{\log{p(x^*_{n,k,d}|\theta{})}}
\end{equation}
Next, we note that the mean (variance) of a sum of independent random variables is simply the sum of the mean (variance) of the independent random variables. Thereafter, it is straightforward to decompose \ExNLLShort and \VarNLLShort into a sum of their components, resulting in two types of explanation methods, 
\begin{equation}
    \Phi^{\E{}_{\theta}}{(\textbf{x}_n^* {;}\:\theta{})} = [\sum_{d=1}^{D}{\E{}_{\theta}{[\minus{\log{p(x^*_{n,k,d}|\theta{})}]}}}]_{k=1}^{K}
\end{equation}
\begin{equation}
    \Phi^{\text{Var}_{\theta}}{(\textbf{x}_n^* {;}\:\theta{})} = [\sum_{d=1}^{D}{\text{Var}_{\theta}{[\minus{\log{p(x^*_{n,k,d}|\theta{})}]}}}]_{k=1}^{K}
\end{equation}
The $\phi_{n,k}^{\E{}_{\theta}}$ and $\phi_{n,k}^{\text{Var}_{\theta}}$ are local explanations for each BAE prediction which have higher granularity than global explanations attained through conventional feature selection methods \cite{CHANDRASHEKAR201416} which assign importance to features based on the entire dataset. One way to obtain such global explanations from local explanations is by averaging the sensor attribution scores over the dataset  $[\:\frac{1}{N}\sum_{n=1}^{N}{\phi_{n,k}}\:]_{k=1}^K$.

Furthermore, we emphasise our analysis on the sensor level which is closer to application, although we can also decompose the HI into the lower feature level. That is, one would expect an operator to be more capable of acting upon the sensors rather than the measurements themselves. 

\subsection{Centralised and Coalitional BAE}

\begin{figure}[H]
     \centering
     \begin{subfigure}[b]{0.8\textwidth}
         \centering
         \includegraphics[width=\textwidth]{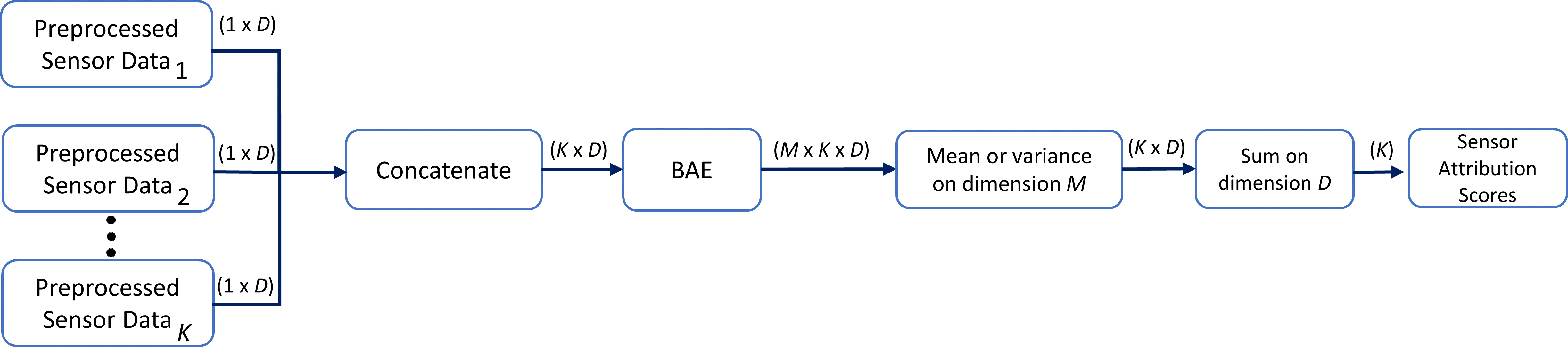}
         \caption{Centralised BAE}
         \label{subfig:centralised}
     \end{subfigure}
     \hfill
     \begin{subfigure}[b]{0.8\textwidth}
         \centering
         \includegraphics[width=\textwidth]{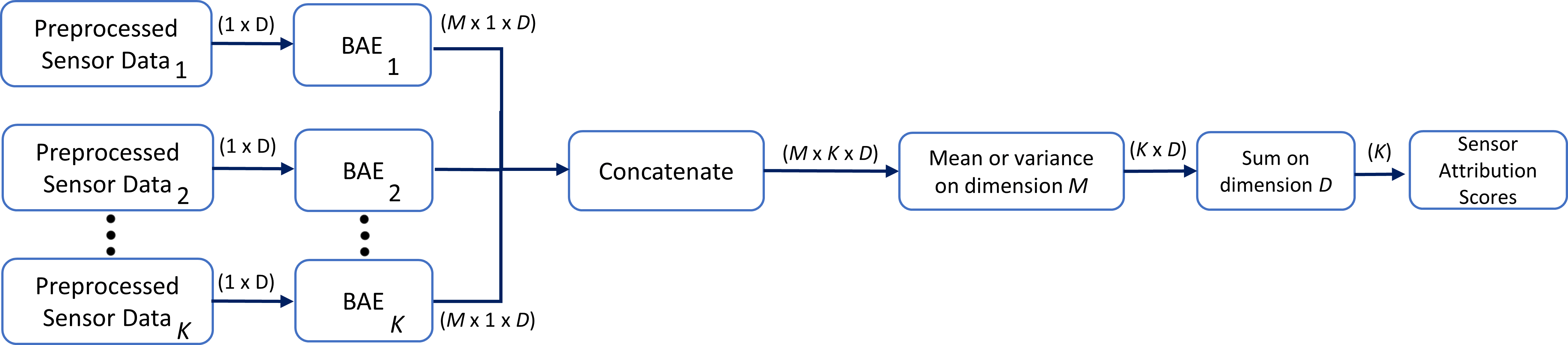}
         \caption{Coalitional BAE}
         \label{subfig:coalitional}
     \end{subfigure}
     \caption{Data flow for the configurations of (a) Centralised BAE and (b) Coalitional BAE with $K$ sensors where each sensor measures $D$ features. The number of BAE samples (size of ensemble) is $M$. }\label{fig:centralised-vs-coalitional-bae}
\end{figure}

In a usual setup of BAE, one would concatenate multiple sensor data streams and forward pass to a single BAE model on the entire data, which we call Centralised BAE  (\cref{fig:centralised-vs-coalitional-bae}(a)). This creates a dependency on all input sensors when computing the attribution score for one sensor (e.g sensor attribution for $s_1$ is dependent on input data from all sensors $s_1, s_2,...,s_K$). As we will find in our empirical results in \cref{section-corr}, this configuration induces correlation in the BAE outputs and subsequently harms the quality of explanation. 

There are two possible sources for the correlation in BAE outputs: (1) correlation in input data and (2) model parameters and architecture. For (1), the correlation in outputs which arise from the correlation in input sensors is actually doing the right thing - explaining the correct state of the correlated sensors. 

Conversely, we posit it is mainly due to (2), that the correlation in BAE outputs are detrimental to the quality of explanations. In this case, neuron activations due to a given sensor may cause neurons for other sensors to be (de)activated. In effect, for a system where one sensor is drifting, this may lead to the Centralised BAE explaining that other sensors are also drifting when in fact, they are not. 

To alleviate this, we propose a scheme of training an independent BAE model for every sensor (\cref{fig:centralised-vs-coalitional-bae}(b)) called Coalitional BAE. By doing so, we force the BAE outputs to be computed independently for each sensor (e.g attribution score for $s_1$ is dependent only on input data from $s_1$ and exclude other sensors). 

One advantage of our proposed configuration is its simplicity of implementation --- no modification in model architecture is required when converting from a centralised setup. During testing phase, we simply concatenate the outputs of the independent BAEs which result in the same dimensions as the Centralised BAE outputs. The concept of a Coalitional BAE is inspired by coalitional agent-based system theory \cite{RAHWAN2007535, rahwan2005distributing} where each sensor is represented by an independent BAE agent. In our context, the BAE agents cooperate in a coalition to achieve a common goal of monitoring the equipment's health. We note that Lundberg and Lee \cite{lundberg2017unified} also leveraged Shapley's value from agent-based system theory to fairly allocate attribution scores to the features. 

Furthermore, our configuration is different from the multi-headed architecture proposed by Canizo et al. \cite{canizo2019multi} which does not guarantee the independence of reconstructed signals since it splits only the early convolution layers while the subsequent layers are fused with recurrent or fully-connected layers.

\subsection{Evaluation Metrics for Sensor Attribution based on Covariate Shift}

An important concept in evaluating the performance of explanation methods is the covariate shift of sensor data in non-stationary environment \cite{raza2014adaptive}. Specifically, we consider covariate shift as the temporal departure of test data from an initial training data distribution (for example, the healthy phase of a manufacturing equipment). Therefore, we emphasise that a good explanation outcome should be reflective of the underlying covariate shift states of the sensors (i.e either shifting or non-shifting). 

Two properties of a sensor attribution score are of relevance; its monotonic relationship with $Y$, and its ability to rank sensor importance. Intuitively, we expect a good explanation method to assign monotonically increasing attribution scores to shifting sensors ($S^{shift}$) and conversely, the attribution scores of non-shifting sensors ($S^{\bot{shift}}$) should be non-monotonic. We also expect these scores to be higher in ranks for $S^{shift}$ when compared to those of $S^{\bot{shift}}$. Hence, we propose metrics to evaluate these desired behaviours of good explanations.

\subsubsection{Notation}
We denote the set of $I$ shifting sensors as $S^{shift} = \{s^{shift}_{i}\}_{i=1}^{I}, S^{shift} \subset S_K$ and $J$ non-shifting sensors as $S^{\bot{shift}} = \{s^{\bot{shift}}_{j}\}_{j=1}^{J=K-I}, S^{\bot{shift}} \subset S_K$ and such that they are disjoint sets $S^{shift} \cap S^{\bot{shift}} = \emptyset$.

Where the unobserved degradation of an equipment is of relevance, $Y = [y_n]_{n=1}^{N}, y_n \in [0,1]$ where $y_n=0$ represents perfect health and $y_n$ increases linearly with $n$ until failure occurs at $y_n=1$. The HI is a model of $Y$ and ideally has a monotonic relationship with $Y$. 

Let $R(\textbf{a}) : \rm I\!R^{dim(\textbf{a})} \to \{r : r \in \mathbb{N}^{+}, r \leq dim(\textbf{a})\}^{dim(\textbf{a})}$ be the ranking function which returns the ranks of elements in an arbitrary vector. For example, $R([0.8,0.1,0.2])=[1,3,2]$. The relation between $i$-th and $j$-th elements of the ranked vectors $R(\textbf{a})_i < R(\textbf{a})_j$ indicates that $a_i$ has a higher rank than $a_j$. 

\subsubsection{G-mean of Spearman drift coefficients, $G_{SDC}$}

The Spearman rank correlation coefficient \cite{spearman1987proof} measures the degree of any monotonic relationship, either linear or non-linear, between two variables. As a non-parametric technique, it does not assume normal distribution in the data and is less sensitive to outliers compared to the Pearson correlation coefficient \cite{gauthier2001detecting}. 
Essentially, the Spearman rank correlation coefficient is the Pearson correlation coefficient which operates on the ranks of variables rather than the raw values. That is, $Spearman(\textbf{a},\textbf{b}) = Pearson(R(\textbf{a}), R(\textbf{b}))$ where $Spearman$ is a function returning the Spearman rank correlation coefficient between any vectors $\textbf{a}$ and $\textbf{b}$, while the $Pearson$ function computes the Pearson correlation coefficient. Next, the Spearman drift coefficient of the $k$-th sensor attribution score can be defined as follows,
\begin{equation}
    \rho{({s_{k}})} = \left\{
        \begin{array}{ll}
            \abs{{Spearman}([\phi_{n,k}]_{n=1}^N, Y)}  & \quad{} ,\: \text{p\minus{}value} \leq \alpha{} \\
            0  & \quad{},\: \text{p\minus{}value} > \alpha{}
        \end{array}
    \right.
\end{equation}
It takes on the absolute value of the Spearman rank correlation coefficient when the p-value is less than or equal to a level of significance $\alpha$ (set as 0.05) and 0 otherwise which implies there is no monotonic relationship. As a balanced evaluation metric, we compute the geometric mean (G-mean) of the Spearman drift coefficients for $S^{shift}$ and $S^{\bot{shift}}$,
\begin{equation}
    G_{SDC} = \sqrt{(\frac{1}{I}\sum_{i=1}^I{{\rho{(s^{shift}_{i})}}}) (\frac{1}{J}\sum_{j=1}^{J}{1-{\rho{(s^{\bot{shift}}_{j})}})}} ,\: G_{SDC} \in [0,1]
\end{equation}
The first term in the square root favours explanation methods which have higher Spearman drift coefficients on $S^{shift}$, while the second term penalises methods which falsely assign high Spearman drift coefficients to $S^{\bot{shift}}$. 

\subsubsection{G-mean of sensitivity-specificity of explanation ranking, $G_{SSER}$}


To evaluate the explanation methods in ranking the sensors importance, now we turn to obtaining ranks of the sensor attribution scores across the set of sensors. Using these set of ranks, we assign positive labels (shifting) to the top-$I$ sensors, and negative labels (non-shifting) otherwise,
\begin{equation}
top\minus{I(r)} = \left\{
        \begin{array}{ll}
            1  & \quad{} ,\: r \leq I\\
            0  & \quad{},\: r > I
        \end{array}
    \right.
\end{equation}
We obtain the assigned binary labels $\textbf{A}=[\:top\minus{I(r)} : r \in R([\phi_{n,k}]_{k=1}^K)\:]_{n=1}^{N}, \textbf{A} \in \{0,1\}^{N\times{K}}$. As one method to obtain \textit{gold} labels, we define the function
\begin{equation}
gold(k) = \left\{
        \begin{array}{ll}
            1  & \quad{} ,\: s_k \in S^{shift}\\
            0  & \quad{},\: s_k \in S^{\bot{shift}} 
        \end{array}
    \right.
\end{equation}
which leads to $\textbf{G} = [\:[\:gold(k)\:]_{k=1}^K\:]_{n=1}^{N}$, $\textbf{G} \in \{0,1\}^{N\times{K}}$. To compare $\textbf{A}$ and $\textbf{G}$, we vectorise them and compute the confusion matrix. Subsequently, as a single performance measure, we calculate the G-mean of the sensitivity-specificity of explanation ranking,
\begin{equation}
    G_{SSER}= \sqrt{\underbrace{\frac{TP}{TP+FN}}_{sensitivity}\,\underbrace{\frac{TN}{TN+FP}}_{specificity}},\: G_{SSER} \in [0,1]
\end{equation}
where $TP$, $TN$, $FP$, and $FN$ are the true positives, true negatives, false positives and false negatives respectively. We prefer the G-mean over the $F_1$-measure since the $F_1$-measure does not account for $TN$ which is relevant in evaluating the ranking of the non-shifting sensors \cite{chicco2020advantages}. Moreover, the G-mean of sensitivity-specificity is a common evaluation metric for imbalanced dataset classification \cite{kuncheva2019instance}. This is relevant in our case, where the number of sensors in $S^{shift}$ can potentially be higher (lower) than $S^{\bot{shift}}$.

\subsubsection{Sensor Explanation Quality Index , $SEQI$}

\begin{figure}[H]
    \centerline{\includegraphics[scale=.65]{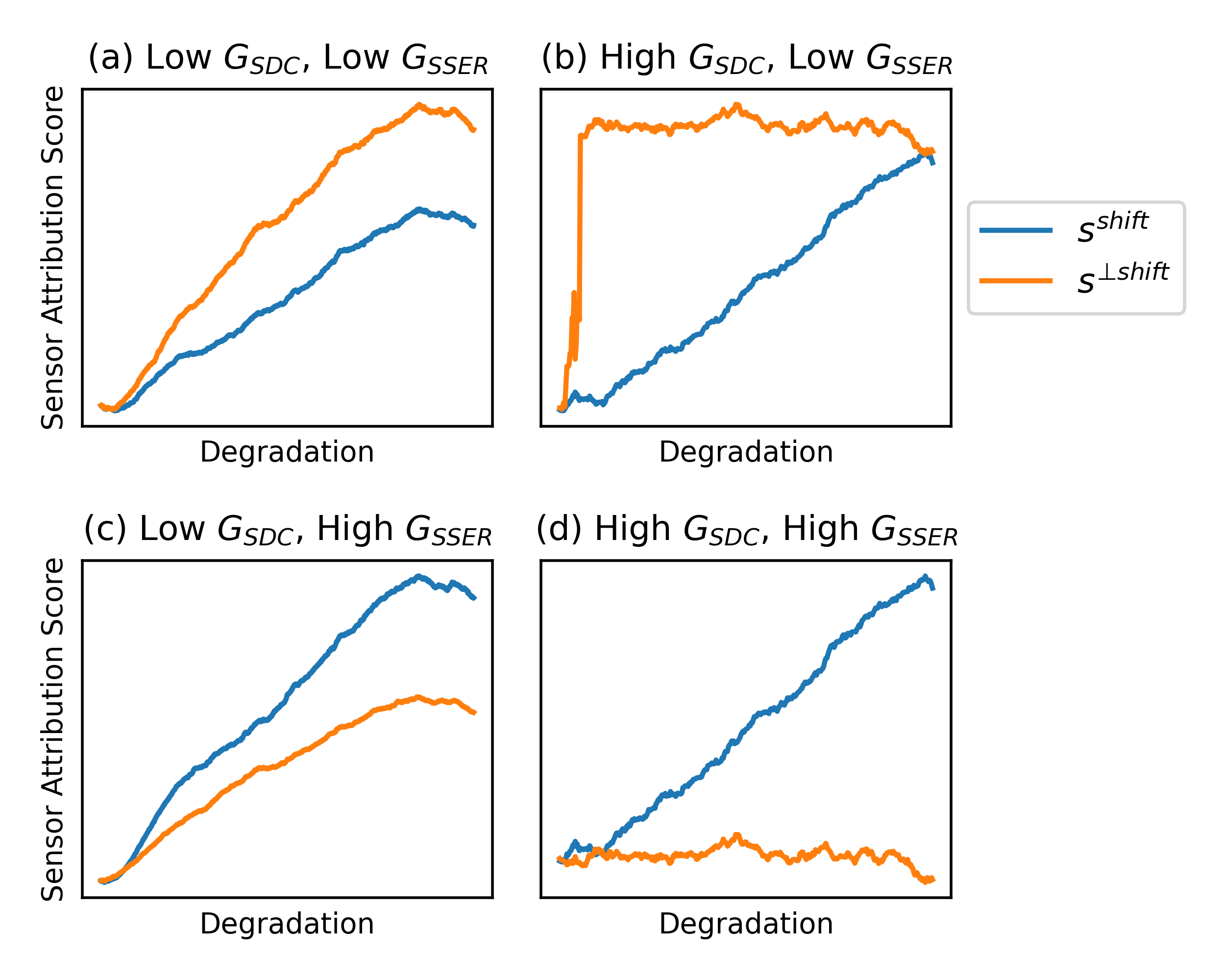}}
    \caption{Non-ideal explanation quality depicted in (a), (b) and (c). Low $G_{SDC}$ is attained when the attribution score for $s^{\bot{shift}}$ has a strong monotonic relationship with equipment degradation in (a) and (c), while explanation methods with poor $G_{SSER}$ are shown in (a) and (b) where the attribution score of $s^{\bot{shift}}$ is consistently higher than $s^{shift}$. (d) depicts the desired explanation quality with high $G_{SDC}$ and $G_{SSER}$. Samples are obtained from evaluation on PRONOSTIA dataset.}
    \label{fig:exp-quality}
\end{figure}

Solely evaluating an explanation method using either $G_{SDC}$ or $G_{SSER}$ is not sufficient; it may perform poorly on either $G_{SDC}$ or $G_{SSER}$ which is not desirable (see \cref{fig:exp-quality}(a), (b) and (c)) while the ideal quality is to have high values on both metrics as depicted in \cref{fig:exp-quality}(d).  

Therefore, we propose to combine $G_{SDC}$ and $G_{SSER}$ as a weighted sum for a more comprehensive evaluation metric which we call Sensor Explanation Quality Index  ($SEQI$), 
\begin{equation}
    SEQI = \omega_1{G_{SDC}} + \omega_2{G_{SSER}},\: SEQI \in [0,1]
\end{equation}
In our study, we weigh both terms equally by setting $\omega_1$ and $\omega_2$ as 0.5. Although our proposed metrics may not capture all possible desiderata of explanations, they can still serve as useful sanity check to compare explanation methods.

\subsubsection{Obtaining labels for test set}

One challenge to compute $G_{SDC}$, $G_{SSER}$ and $SEQI$ is in obtaining temporal datasets with gold labels of which subset of sensors are shifting and non-shifting. We describe a simple preprocessing procedure to augment these labels from existing datasets.

Firstly, we designate a subset of sensors as \DriftSS and the remaining as \nonDriftSS. Then, it is straightforward to obtain the measurements for \DriftSS by directly using the natural data which arise from a non-stationary environment (e.g equipment run-to-failure). Meanwhile for the \nonDriftSS, we employ a bootstrap sampling from the training set to maintain an equal number of data points as the original test set. This step ensures data stationarity for the \nonDriftSS. Using the resampled test set and gold labels, we can proceed to computing the evaluation metrics $G_{SDC}$, $G_{SSER}$ and $SEQI$. To have a more comprehensive evaluation, we should repeat the evaluation procedures by designating a different combination of shifting and \nonDriftSS.

One important outcome of the resampling procedure is that it enforces the \nonDriftSS raw inputs to be uncorrelated with the \DriftSS. This does not enforce, however, the outputs of BAE to be uncorrelated.

\section{Experimental setup} \label{section-exp-setup}

This section describes the experiment setup necessary to reproduce our results of comparing various explanation methods and BAE configurations. 

\subsection{Dataset} \label{subsection-dataset}

We experimented with the proposed methods on two public datasets: PRONOSTIA \cite{nectoux2012pronostia} and ZEMA \cite{zemadataset2019, helwig2017integrated} which consist of sensor measurements collected from multiple equipments' run-to-failure. Each trajectory in these datasets is prepared in an array with the following dimensions: ($N$ cycles $\times$ $K$ sensors $\times$ $D$ measurements).

For the PRONOSTIA dataset, the data were collected on a bearing testbed under three different load conditions with multiple repetitions resulting in a total of 17 trajectories. There were 2 accelerometers (1 horizontal and 1 vertical), each of 25.6kHz sampling frequency. Every cycle had an interval of 10 seconds and each sensor recorded 2560 measurements within a time window which lasted 1/10 of a second. 

For the ZEMA dataset, 3 run-to-failure repetitions were conducted on a electromechanical cylinder (EMC) testbed. A total of 11 sensors consisting of a microphone, 3 vibration sensors mounted on the plain bearing, piston rod, and ball bearing respectively, 1 axial force sensor, 1 pressure sensor, 1 velocity sensor, 1 active current sensor, and 3 motor current phase sensors were installed. In the provided dataset, only the last 1 second of return stroke is available and all measurements were resampled to 2kHz. 

\subsection{BAE Architecture} 
\label{subsection-architecture}

We implemented the BAE using the \textit{baetorch} package\cite{yong2020baetorch} with the parameters and layers stated in \cref{table:architecture}. The size of the ensemble was set to 5. For the Centralised BAE, the input channel size of the first convolution layer is set to the number of sensors present in each dataset. Similar architecture was used for each BAE agent in the Coalitional BAE setup while fixing the input channel size of the first convolution layer to 1. 
\begin{table}[H]
\caption{Encoder architecture of BAE for PRONOSTIA and ZEMA datasets. The decoder is a reflection of the encoder without the dense latent layer, and the convolution layers are replaced by convolution transpose layers.}
\label{table:architecture}
\centering
\begin{tabular}{@{}cccc@{}}
\toprule
Layer       & Output channels/nodes & Kernel size & Strides \\ \midrule
1D-Convolution & 8      & 20  & 4   \\
1D-Convolution & 5      & 10  & 2   \\
1D-Convolution & 3      & 5  & 1  \\
Flatten     & -              & -      & -       \\
Dense       & 100       & -      & -      \\
Dense (latent)      & 20       & -      & -      \\ \bottomrule

\end{tabular}
\end{table}
The leaky ReLu \cite{maas2013rectifier} was used as the activation function for each layer with a slope of 0.01. The sigmoid function was applied to scale each reconstructed signal to [0,1].

\subsection{Data pipeline}

With the datasets and BAE architecture in \cref{subsection-dataset} and \cref{subsection-architecture}, we implemented the data pipeline using \textit{agentMET4FOF} package \cite{yong2020agentMET4FOF} for managing the data flow and model classes in a modular and reproducible way (\cref{appendix:agents}). 

The data pipeline is executed repeatedly using various hyperparameters, explanation methods and model configurations. We refer the model capacity as the number of output channels and nodes of each layer, and the kernel size of the convolution layers. The data pipeline has the BAE model capacity evaluated at three levels: x$\frac{1}{2}$, x1 and x2 for both Centralised and Coalitional BAEs. At each model capacity level, we vary the depth of the BAE by setting the number of convolution layers at three levels: 1 layer, 2 layers and 3 layers. 

We also repeat the experiment runs by replacing the BAE with a deterministic AE (size of ensemble as 1). For comparison against posthoc explanation methods, we apply DeepLift \cite{shrikumar2017learning} and GradientSHAP \cite{lundberg2017unified} on the deterministic AE implemented using the \textit{Captum} package\footnote{https://captum.ai} to obtain sensor attribution scores for the reconstruction loss. The data pipeline is illustrated in \cref{fig:overall-pipeline} and described as follows:

\begin{figure}[H]
\centerline{\includegraphics[scale=.60]{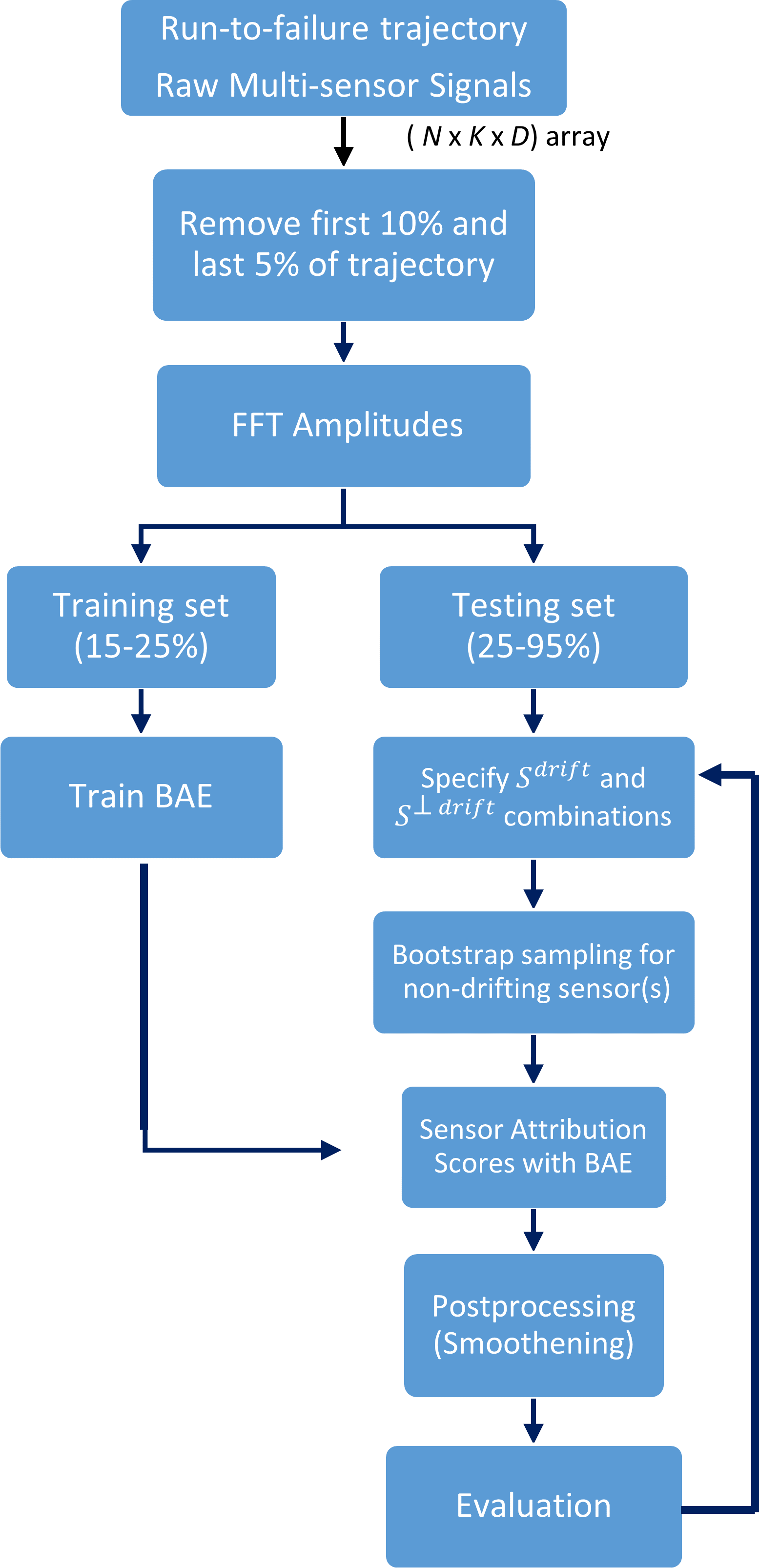}}
\caption{Data pipeline for evaluating BAE explanation quality}
\label{fig:overall-pipeline}
\end{figure}

\begin{enumerate}
    \item The raw multi-sensor data in the run-to-failure trajectory was preprocessed in the form of array with dimensions of ($N$ cycles $\times$ $K$ sensors $\times$ $D$ measurements).
    
    \item We removed the first 10\% and last 5\% of the cycles which correspond to the run-in and total failure phases respectively. This is because the measurements during these phases can be highly transient and unstable. Moreover, it is of less practical value for a HI which is too near the point of failure. 
    
    \item We applied fast Fourier transform \cite{brigham1967fast} to obtain the absolute amplitudes in the frequency spectrum. This reduces the total feature size by half.
    
    \item Without shuffling the data, we split the trajectory and use the first 15-25\% of the cycles for training and remaining 25-95\% for testing. 
    
    \item During the training phase, we fitted a Min-Max scaler \cite{scikit-learn} on the training set to rescale the feature values independently to [0,1]. Then, we trained the BAE on the rescaled data using either the centralised or coalitional configuration. We ran an automatic learning rate finder \cite{smith2017cyclical} and minimised the loss in \cref{eqn:bae-loss} using Adam optimiser \cite{kingma2014adam} with $\lambda{}$ set to 0.001. The training lasted for 250 epochs.
    
    \item For the test set, we firstly decided on a subset of sensors to be labelled as shifting ($S^{shift}$) while the remaining sensors were labelled as non-shifting ($S^{\bot{shift}}$). Data for $S^{\bot{shift}}$ was bootstrap sampled from the training set to have equal number of data points as the test set which ensures the data is always from the training distribution. The $S^{shift}$ have their data unchanged from the test set.
    
    \item After preparing the test set, we fed the data to the Min-Max scaler and the BAE. We obtained the sensor attribution scores using one of the explanation methods. 
    
    \item As a postprocessing step, simple moving average was applied on the sensor attribution scores with a window size equal to the total cycles in the training dataset. We found it  helpful to reduce fluctuations from cycle-to-cycle for a more stable result. For non-posthoc explanation methods such as \ExNLLShort and \VarNLLShort, the attribution scores were further normalised to begin from 1 by dividing them by the mean values computed from the training set.  
    
    \item The $G_{SDC}$, $G_{SSER}$ , $SEQI$ are evaluated. To investigate the correlation in explanations, we also calculated the Pearson correlation coefficient between the sensor attribution scores of $S^{shift}$ and $S^{\bot{shift}}$. For ZEMA, we took the sensor attribution scores averaged across $S^{shift}$ and $S^{\bot{shift}}$.
    
    \item Steps 6 to 9 were repeated with varying combinations of $S^{shift}$ and $S^{\bot{shift}}$. For PRONOSTIA, we repeated using all possible combinations for 2 sensors. For ZEMA, due to the large number of all possible combinations for 11 sensors, we evaluated using combinations of 1 and 10 shifting sensors resulting in 22 combinations. 

    \item Upon completing all evaluations on the combinations of $S^{shift}$ and $S^{\bot{shift}}$, we repeat steps 1 to 10 with another run-to-failure trajectory from the PRONOSTIA or ZEMA datasets. 
        
\end{enumerate}

\section{Results and Discussion} \label{section-res-dis}

We present our results of investigating the presence of correlation in the sensor attribution scores. Then, we move on to a comprehensive comparison between various BAE configurations and explanation methods based on the proposed metrics of $G_{SDC}$, $G_{SSER}$ and $SEQI$. Additional results are shown in \cref{appendix:additional}.

\subsection{Correlation in BAE's explanations} \label{section-corr}

\begin{figure}[H]
    \begin{subfigure}{\textwidth}
      \centering
      \caption{PRONOSTIA}
      \includegraphics[width=\textwidth]{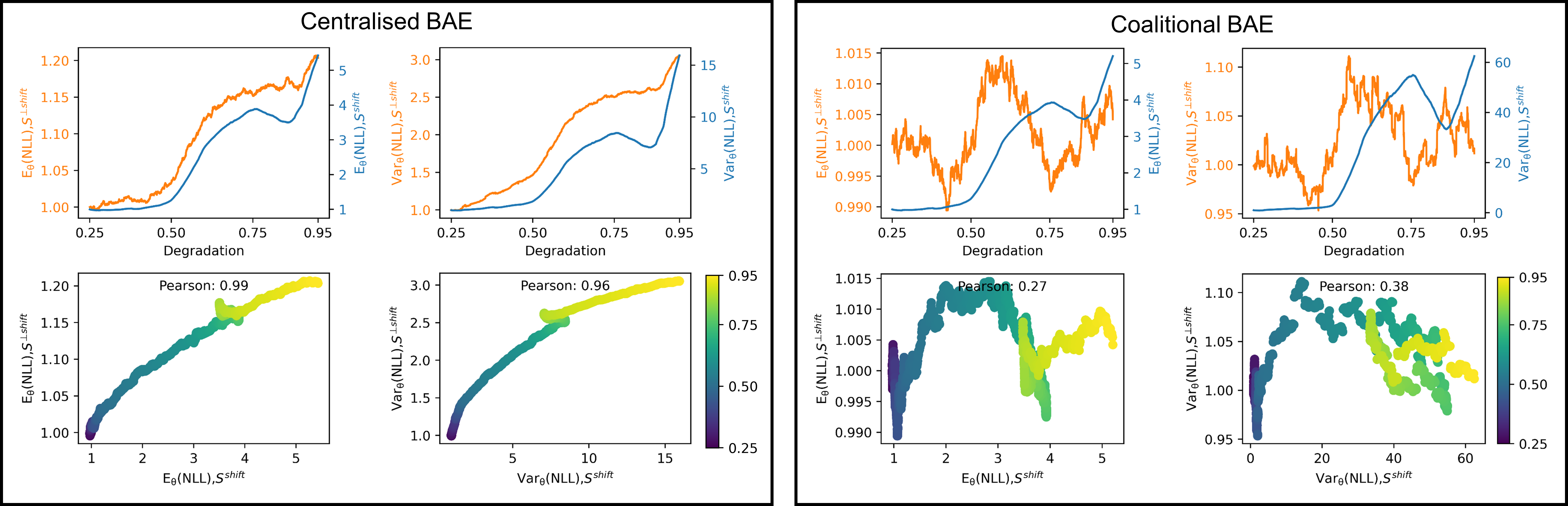}
      
    \end{subfigure}
    \begin{subfigure}{\textwidth}
      \centering
      \caption{ZEMA}
      \includegraphics[width=\textwidth]{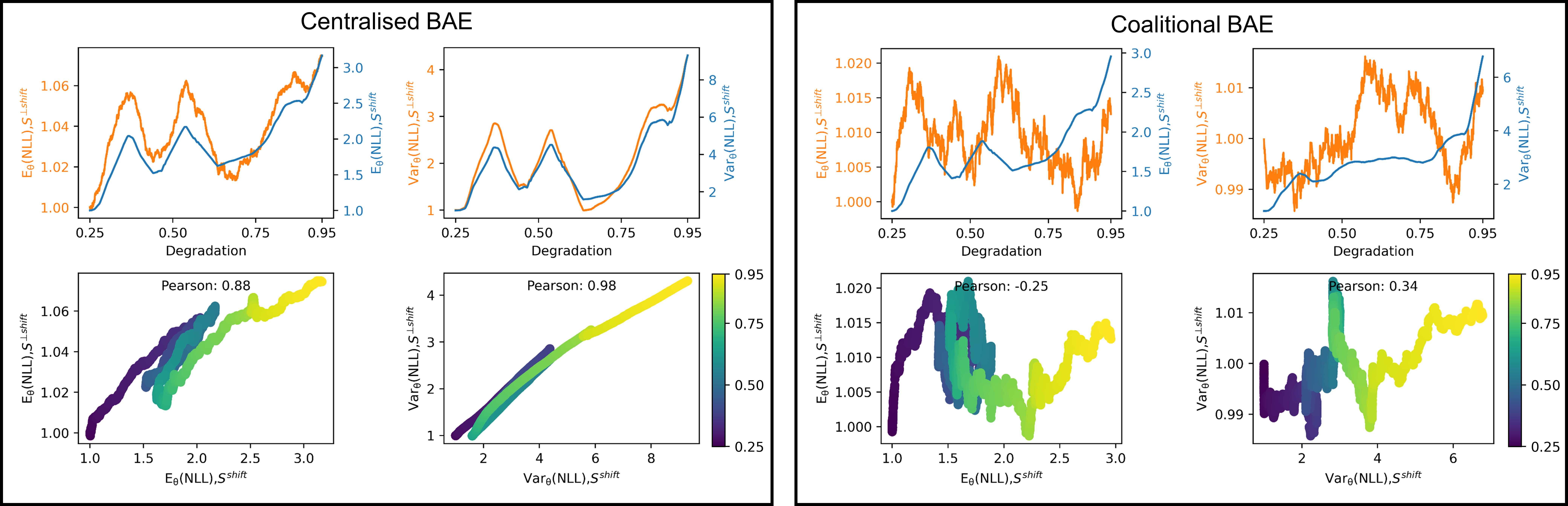}
      
    \end{subfigure}
    \caption{Samples from (a) PRONOSTIA and (b) ZEMA datasets showing the attribution scores for \DriftSS and \nonDriftSS obtained with the Centralised and Coalitional BAEs using the \ExNLLShort and \VarNLLShort explanation methods. Top and bottom rows in each panel are different depictions of the same data. Color bar of bottom row indicate the equipment degradation. The values of Pearson correlation coefficients are also shown in the bottom row.}
    \label{fig:corr-samples}
\end{figure}

Surprisingly, from \cref{fig:corr-samples}, we observe that the Centralised BAE induces strong correlation between its explanations : the attribution scores for \DriftSS take on similar shape as \nonDriftSS and the Pearson correlation coefficient is high ($\geq{0.80}$) on samples from both PRONOSTIA and ZEMA. This applies to both \ExNLLShort and \VarNLLShort methods. This observation is crucial, it implies these attribution scores can yield misleading explanations that some sensors are monotonically shifting away from their healthy data distribution when in fact, they are not. It was due to this observation that we proposed the Coalitional BAE which enforces independence in the sensor attribution scores, thereby reducing the correlation dramatically. 

Another observation is that the absolute values of attribution scores for \nonDriftSS are also increased for Centralised BAE due to the correlation, and are higher than those of Coalitional BAE. As such, with the smaller differences between attribution scores for \DriftSS and \nonDriftSS, the ability to rank sensor importance may also be poorer with the use of the centralised configuration. 

\begin{figure}[H]
\centering
\includegraphics[width=\linewidth]{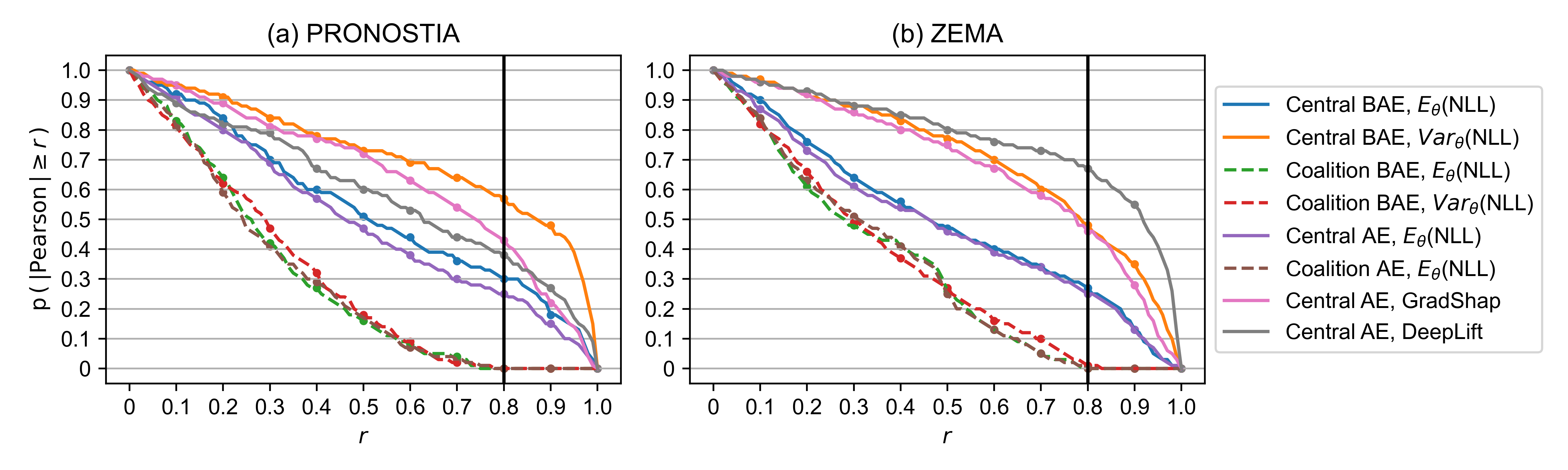}
\caption{Complementary empirical distribution function of the Pearson correlation coefficient between sensor attribution scores of \DriftSS and \nonDriftSS for (a) PRONOSTIA and (b) ZEMA datasets for various BAE configurations and explanation methods. Thick vertical line is drawn at r=0.8 which is deemed as high Pearson correlation coefficient threshold. The empirical probability is evaluated over all experimental runs.}
\label{fig:pcorr}
\end{figure}

We further investigate this phenomenon with the complementary empirical distribution function depicted in \cref{fig:pcorr}. It is confirmed that the centralised configuration (solid lines) has consistently greater chances of exhibiting high correlation between the attribution scores for \DriftSS and \nonDriftSS than the coalitional configuration (dotted lines). 

Strikingly, the Centralised BAE with \VarNLLShort explanations have more than half of the samples with high correlation on both PRONOSTIA and ZEMA. In contrast, the Coalitional BAE counterpart has very close to zero sample with high correlation. Furthermore, applying the posthoc explanation methods DeepLift and GradientSHAP on the Centralised AE also exhibit high chances of strong correlation in their attribution scores.

\subsection{Quantitative evaluation of explanation quality}

\begin{figure}[H]
    \begin{subfigure}{\textwidth}
      \centering
      \includegraphics[width=\linewidth]{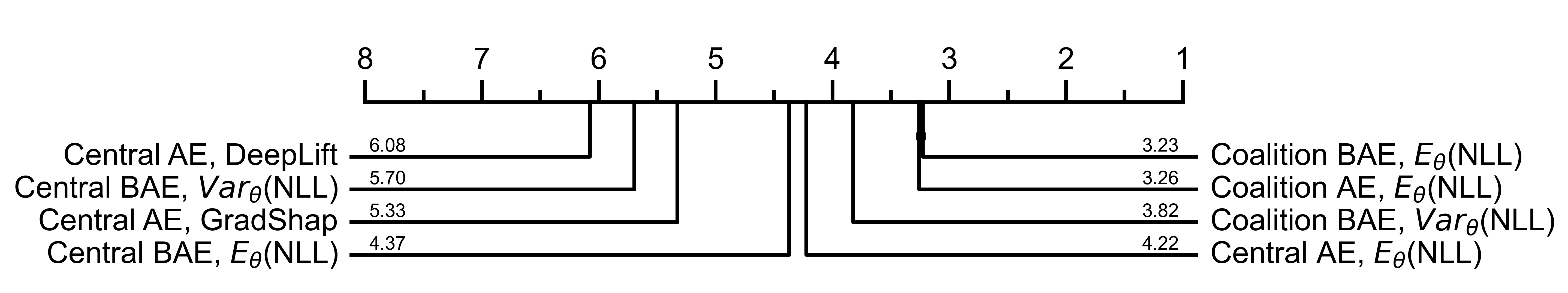}  
      \caption{$G_{SDC}$}
      \label{subfig:cd-gsdc}
    \end{subfigure}
    \begin{subfigure}{\textwidth}
      \centering
      \includegraphics[width=\linewidth]{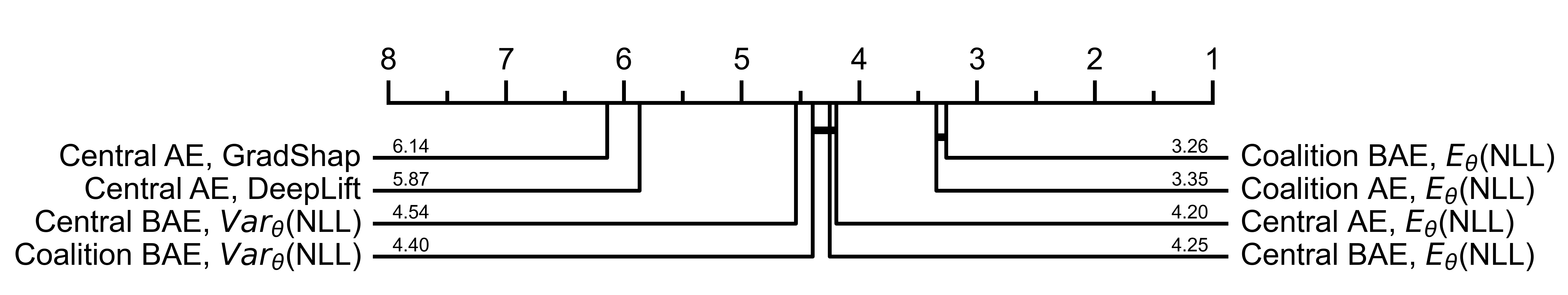}
      \caption{$G_{SSER}$}
      \label{subfig:cd-sser}
    \end{subfigure}
        \begin{subfigure}{\textwidth}
      \centering
      \includegraphics[width=\linewidth]{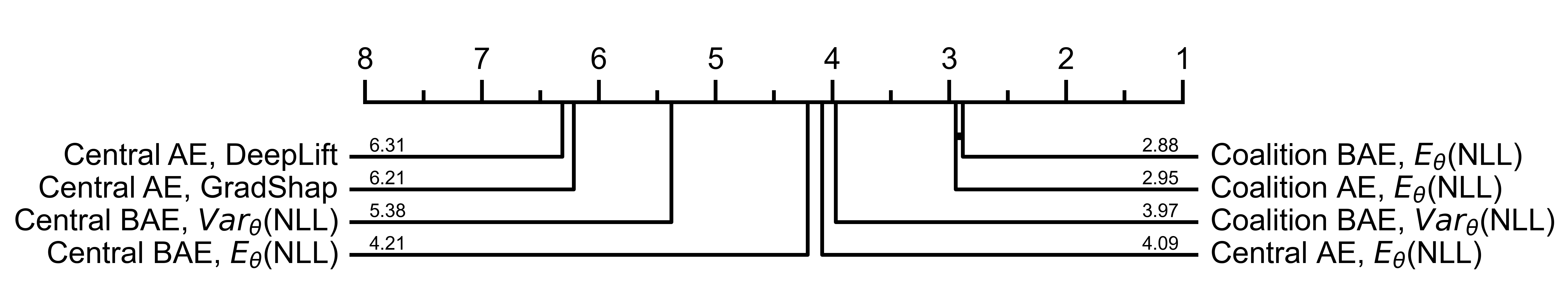}
      \caption{$SEQI$}
      \label{subfig:cd-seqi}
    \end{subfigure}
    \caption{Critical difference diagrams for comparing explanation methods under evaluation metrics of (a) $G_{SDC}$, (b) $G_{SSER}$ and (c) $SEQI$. Average rank of each method is shown next to the labels. The methods on the right have higher ranks than the methods on left. A thick line connecting the methods indicate there is no significant difference between the methods.}
    \label{fig:main-cd-diagrams}
\end{figure}

We tabulate the results of our experiments (\cref{table:pronostia-gsdc}, \cref{table:pronostia-sser}, \cref{table:pronostia-seqi}) and further analyse the differences between various BAE configurations and explanation methods by visualising the critical difference diagrams (\cref{fig:main-cd-diagrams}) which are commonly used to compare machine learning models' performances across multiple datasets and hyperparameter settings \cite{demvsar2006statistical, IsmailFawaz2018deep}. Essentially, the critical difference diagram reports the results of a pairwise posthoc analysis. For this, we adopt the non-parametric Wilcoxon signed-rank test \cite{wilcoxon1992individual} after failing to reject the null hypothesis of a Friedman test \cite{friedman1940comparison}. 

\begin{table}[H]
\centering
\caption{Mean and standard error of $G_{SDC}$ (0 = worst score, 1 = perfect score) for various explanation methods with varied number of convolution layers and model capacities evaluated on (a) PRONOSTIA and (b) ZEMA datasets. Highest mean value of each column is bolded. * indicates overall highest value. }
\scriptsize  (a) PRONOSTIA

\label{table:pronostia-gsdc}
\resizebox{\textwidth}{!}{
\begin{tabular}{llllllllllll}
\hline
&  &  & \multicolumn{9}{c}{G-mean of Spearman drift coefficients ($G_{SSC}$)} \\ \cline{4-12} 
&  &  & \multicolumn{3}{c}{1 Convolution Layer} & \multicolumn{3}{c}{2 Convolution Layers} & \multicolumn{3}{c}{3 Convolution Layers}  \\ \cline{4-12} 

    


\multicolumn{3}{c}{Method \textbackslash{ Capacity }} & \multicolumn{1}{c}{x$\frac{1}{2}$} & \multicolumn{1}{c}{x$1$} &
\multicolumn{1}{c}{x$2$} & \multicolumn{1}{c}{x$\frac{1}{2}$} & \multicolumn{1}{c}{x$1$} & \multicolumn{1}{c}{x$2$} & \multicolumn{1}{c}{x$\frac{1}{2}$} & \multicolumn{1}{c}{x$1$} & \multicolumn{1}{c}{x$2$} \\ \hline


AE  & Central   & \exLL{} &    
0.606$\pm{}$0.038&0.566$\pm{}$0.035&0.584$\pm{}$0.039&0.594$\pm{}$0.038&0.547$\pm{}$0.041&0.552$\pm{}$0.037&0.653$\pm{}$0.036&0.633$\pm{}$0.039&0.573$\pm{}$0.043

\\ \cline{3-12}  &           & DeepLIFT & 
0.500$\pm{}$0.042&0.419$\pm{}$0.035&0.396$\pm{}$0.040&0.487$\pm{}$0.048&0.366$\pm{}$0.044&0.407$\pm{}$0.040&0.537$\pm{}$0.043&0.452$\pm{}$0.035&0.386$\pm{}$0.040

\\ \cline{3-12}  &           & GradShap & 
0.558$\pm{}$0.030&0.522$\pm{}$0.034&0.531$\pm{}$0.030&0.529$\pm{}$0.031&0.511$\pm{}$0.023&0.481$\pm{}$0.030&0.530$\pm{}$0.026&0.489$\pm{}$0.028&0.487$\pm{}$0.032

\\ \cline{2-12}  & Coalition & \exLL{}     & 
\textBF{0.714$\pm{}$0.037}&\textBF{*0.742$\pm{}$0.031}&0.715$\pm{}$0.028&0.668$\pm{}$0.034&0.694$\pm{}$0.031&\textBF{0.698$\pm{}$0.033}&0.682$\pm{}$0.032&0.661$\pm{}$0.043&\textBF{0.707$\pm{}$0.030}

\\ \hline BAE & Central   & \exLL{}     &
0.527$\pm{}$0.037&0.527$\pm{}$0.032&0.557$\pm{}$0.037&0.574$\pm{}$0.037&0.560$\pm{}$0.041&0.525$\pm{}$0.036&0.676$\pm{}$0.035&0.598$\pm{}$0.041&0.603$\pm{}$0.038
  
\\ \cline{3-12}   &           & \varLL{}    &
0.456$\pm{}$0.037&0.449$\pm{}$0.033&0.467$\pm{}$0.038&0.370$\pm{}$0.040&0.425$\pm{}$0.045&0.336$\pm{}$0.044&0.701$\pm{}$0.040&0.404$\pm{}$0.041&0.358$\pm{}$0.044

\\ \cline{2-12}  & Coalition & \exLL{}     & 
0.682$\pm{}$0.038&0.716$\pm{}$0.029&\textBF{0.719$\pm{}$0.030}&0.638$\pm{}$0.041&0.660$\pm{}$0.038&0.680$\pm{}$0.039&0.673$\pm{}$0.033&0.691$\pm{}$0.033&0.696$\pm{}$0.032

\\ \cline{3-12}  &  & \varLL{}   &
0.649$\pm{}$0.041&0.630$\pm{}$0.036&0.613$\pm{}$0.035&\textBF{0.712$\pm{}$0.034}&\textBF{0.696$\pm{}$0.032}&0.655$\pm{}$0.037&\textBF{0.705$\pm{}$0.036}&\textBF{0.726$\pm{}$0.034}&0.685$\pm{}$0.033

    \\ \hline
\end{tabular}
}
\end{table}
\begin{table}[H]
\centering
\scriptsize  (b) ZEMA
\resizebox{\textwidth}{!}{
\begin{tabular}{llllllllllll}
\hline
&  &  & \multicolumn{9}{c}{G-mean of Spearman drift coefficients ($G_{SDC}$)} \\ \cline{4-12} 
&  &  & \multicolumn{3}{c}{1 Convolution Layer} & \multicolumn{3}{c}{2 Convolution Layers} & \multicolumn{3}{c}{3 Convolution Layers}  \\ \cline{4-12} 
    


\multicolumn{3}{c}{Method \textbackslash{ Capacity }} & \multicolumn{1}{c}{x$\frac{1}{2}$} & \multicolumn{1}{c}{x$1$} &
\multicolumn{1}{c}{x$2$} & \multicolumn{1}{c}{x$\frac{1}{2}$} & \multicolumn{1}{c}{x$1$} & \multicolumn{1}{c}{x$2$} & \multicolumn{1}{c}{x$\frac{1}{2}$} & \multicolumn{1}{c}{x$1$} & \multicolumn{1}{c}{x$2$} \\ \hline


AE  & Central   & \exLL{} &    
0.643$\pm{}$0.023&0.620$\pm{}$0.023&0.595$\pm{}$0.025&0.604$\pm{}$0.025&0.619$\pm{}$0.026&0.574$\pm{}$0.025&0.624$\pm{}$0.024&0.587$\pm{}$0.025&0.583$\pm{}$0.024

\\ \cline{3-12}  &           & DeepLIFT & 
0.436$\pm{}$0.022&0.454$\pm{}$0.025&0.436$\pm{}$0.026&0.478$\pm{}$0.022&0.462$\pm{}$0.025&0.442$\pm{}$0.025&0.562$\pm{}$0.029&0.474$\pm{}$0.025&0.501$\pm{}$0.024

\\ \cline{3-12}  &           & GradShap & 
0.490$\pm{}$0.020&0.458$\pm{}$0.021&0.454$\pm{}$0.023&0.540$\pm{}$0.022&0.570$\pm{}$0.022&0.486$\pm{}$0.025&0.578$\pm{}$0.026&0.526$\pm{}$0.022&0.567$\pm{}$0.024

\\ \cline{2-12}  & Coalition & \exLL{}     & 
0.648$\pm{}$0.024&0.659$\pm{}$0.025&0.672$\pm{}$0.024&\textBF{0.659$\pm{}$0.021}&0.647$\pm{}$0.025&0.640$\pm{}$0.024&\textBF{0.654$\pm{}$0.021}&0.646$\pm{}$0.022&0.627$\pm{}$0.027

\\ \hline BAE & Central   & \exLL{}     &
0.638$\pm{}$0.023&0.604$\pm{}$0.025&0.578$\pm{}$0.026&0.615$\pm{}$0.024&0.597$\pm{}$0.025&0.548$\pm{}$0.026&0.644$\pm{}$0.024&0.592$\pm{}$0.024&0.600$\pm{}$0.024
   
\\ \cline{3-12}   &           & \varLL{}    &
0.560$\pm{}$0.022&0.498$\pm{}$0.020&0.473$\pm{}$0.019&0.533$\pm{}$0.017&0.505$\pm{}$0.021&0.493$\pm{}$0.021&0.566$\pm{}$0.018&0.516$\pm{}$0.015&0.528$\pm{}$0.020

\\ \cline{2-12}  & Coalition & \exLL{}     & 
0.659$\pm{}$0.024&\textBF{0.675$\pm{}$0.026}&\textBF{0.675$\pm{}$0.024}&0.653$\pm{}$0.023&0.648$\pm{}$0.024&\textBF{0.647$\pm{}$0.023}&0.643$\pm{}$0.023&\textBF{0.648$\pm{}$0.023}&\textBF{0.632$\pm{}$0.023}

\\ \cline{3-12}  &  & \varLL{}   &
\textBF{*0.680$\pm{}$0.019}&0.621$\pm{}$0.020&0.650$\pm{}$0.020&0.645$\pm{}$0.019&\textBF{0.660$\pm{}$0.019}&0.620$\pm{}$0.021&0.595$\pm{}$0.019&0.631$\pm{}$0.020&0.631$\pm{}$0.020

    \\ \hline
\end{tabular}
}
\end{table}
\begin{table}[H]
\centering
\caption{Mean and standard error of $G_{SSER}$ (0 = worst score, 1 = perfect score) for various explanation methods with varied number of convolution layers and model capacities evaluated on (a) PRONOSTIA and (b) ZEMA datasets. Highest mean value of each column is bolded. * indicates overall highest value. }
\scriptsize  (a) PRONOSTIA
\label{table:pronostia-sser}
\resizebox{\textwidth}{!}{
\begin{tabular}{llllllllllll}
\hline
&  &  & \multicolumn{9}{c}{G-mean of sensitivity-specificity of explanation ranking ($G_{SSER}$)} \\ \cline{4-12} 
&  &  & \multicolumn{3}{c}{1 Convolution Layer} & \multicolumn{3}{c}{2 Convolution Layers} & \multicolumn{3}{c}{3 Convolution Layers}  \\ \cline{4-12} 

    


\multicolumn{3}{c}{Method \textbackslash{ Capacity }} & \multicolumn{1}{c}{x$\frac{1}{2}$} & \multicolumn{1}{c}{x$1$} &
\multicolumn{1}{c}{x$2$} & \multicolumn{1}{c}{x$\frac{1}{2}$} & \multicolumn{1}{c}{x$1$} & \multicolumn{1}{c}{x$2$} & \multicolumn{1}{c}{x$\frac{1}{2}$} & \multicolumn{1}{c}{x$1$} & \multicolumn{1}{c}{x$2$} \\ \hline


AE  & Central   & \exLL{} &    
0.769$\pm{}$0.053&0.826$\pm{}$0.051&0.933$\pm{}$0.031&0.621$\pm{}$0.058&0.692$\pm{}$0.057&0.681$\pm{}$0.056&0.589$\pm{}$0.059&0.629$\pm{}$0.053&0.628$\pm{}$0.056

\\ \cline{3-12}  &           & DeepLIFT & 
0.496$\pm{}$0.064&0.585$\pm{}$0.063&0.651$\pm{}$0.065&0.480$\pm{}$0.064&0.475$\pm{}$0.063&0.495$\pm{}$0.066&0.490$\pm{}$0.066&0.494$\pm{}$0.064&0.497$\pm{}$0.065

\\ \cline{3-12}  &           & GradShap & 
0.610$\pm{}$0.080&0.655$\pm{}$0.075&0.737$\pm{}$0.067&0.704$\pm{}$0.066&0.687$\pm{}$0.068&0.697$\pm{}$0.067&0.660$\pm{}$0.062&0.655$\pm{}$0.064&0.729$\pm{}$0.064

\\ \cline{2-12}  & Coalition & \exLL{}     & 
\textBF{0.818$\pm{}$0.050}&0.883$\pm{}$0.038&0.977$\pm{}$0.010&0.600$\pm{}$0.059&0.713$\pm{}$0.060&0.718$\pm{}$0.059&0.597$\pm{}$0.057&0.624$\pm{}$0.056&0.691$\pm{}$0.052

\\ \hline BAE & Central   & \exLL{}     &
0.754$\pm{}$0.057&0.826$\pm{}$0.049&0.874$\pm{}$0.041&0.604$\pm{}$0.060&0.679$\pm{}$0.059&0.661$\pm{}$0.058&0.606$\pm{}$0.056&0.610$\pm{}$0.058&0.639$\pm{}$0.057

\\ \cline{3-12}   &           & \varLL{}    &
0.767$\pm{}$0.057&0.800$\pm{}$0.058&0.856$\pm{}$0.044&0.605$\pm{}$0.064&0.677$\pm{}$0.062&0.652$\pm{}$0.059&\textBF{0.727$\pm{}$0.055}&0.692$\pm{}$0.061&0.637$\pm{}$0.060

\\ \cline{2-12}  & Coalition & \exLL{}     & 
0.808$\pm{}$0.053&\textBF{0.917$\pm{}$0.035}&\textBF{*0.985$\pm{}$0.008}&0.654$\pm{}$0.061&0.681$\pm{}$0.059&\textBF{0.725$\pm{}$0.059}&0.593$\pm{}$0.059&0.679$\pm{}$0.058&0.669$\pm{}$0.055

\\ \cline{3-12}  &  & \varLL{}   &
0.641$\pm{}$0.061&0.584$\pm{}$0.065&0.679$\pm{}$0.067&\textBF{0.821$\pm{}$0.040}&\textBF{0.728$\pm{}$0.053}&0.683$\pm{}$0.063&0.713$\pm{}$0.061&\textBF{0.827$\pm{}$0.048}&\textBF{0.797$\pm{}$0.048}

    \\ \hline
\end{tabular}
}
\end{table}
\begin{table}[H]
\centering
\scriptsize  (b) ZEMA
\resizebox{\textwidth}{!}{
\begin{tabular}{llllllllllll}
\hline
&  &  & \multicolumn{9}{c}{G-mean of sensitivity-specificity of explanation ranking ($G_{SSER}$)} \\ \cline{4-12} 
&  &  & \multicolumn{3}{c}{1 Convolution Layer} & \multicolumn{3}{c}{2 Convolution Layers} & \multicolumn{3}{c}{3 Convolution Layers}  \\ \cline{4-12} 
    
    


\multicolumn{3}{c}{Method \textbackslash{ Capacity }} & \multicolumn{1}{c}{x$\frac{1}{2}$} & \multicolumn{1}{c}{x$1$} &
\multicolumn{1}{c}{x$2$} & \multicolumn{1}{c}{x$\frac{1}{2}$} & \multicolumn{1}{c}{x$1$} & \multicolumn{1}{c}{x$2$} & \multicolumn{1}{c}{x$\frac{1}{2}$} & \multicolumn{1}{c}{x$1$} & \multicolumn{1}{c}{x$2$} \\ \hline


AE  & Central   & \exLL{} &    
0.589$\pm{}$0.043&0.591$\pm{}$0.042&0.623$\pm{}$0.040&0.590$\pm{}$0.044&0.599$\pm{}$0.041&0.585$\pm{}$0.043&0.606$\pm{}$0.043&0.558$\pm{}$0.046&0.565$\pm{}$0.042

\\ \cline{3-12}  &           & DeepLIFT & 
0.489$\pm{}$0.047&0.404$\pm{}$0.048&0.399$\pm{}$0.048&0.476$\pm{}$0.047&0.483$\pm{}$0.048&0.379$\pm{}$0.049&0.366$\pm{}$0.047&0.437$\pm{}$0.046&0.433$\pm{}$0.049

\\ \cline{3-12}  &           & GradShap & 
0.151$\pm{}$0.041&0.142$\pm{}$0.040&0.157$\pm{}$0.041&0.173$\pm{}$0.042&0.176$\pm{}$0.042&0.173$\pm{}$0.042&0.217$\pm{}$0.048&0.169$\pm{}$0.042&0.203$\pm{}$0.047

\\ \cline{2-12}  & Coalition & \exLL{}     & 
0.677$\pm{}$0.035&0.719$\pm{}$0.031&\textBF{0.688$\pm{}$0.034}&0.647$\pm{}$0.038&0.679$\pm{}$0.034&0.654$\pm{}$0.037&0.635$\pm{}$0.041&\textBF{0.672$\pm{}$0.035}&0.622$\pm{}$0.041

\\ \hline BAE & Central   & \exLL{}     &
0.590$\pm{}$0.043&0.613$\pm{}$0.041&0.624$\pm{}$0.040&0.598$\pm{}$0.040&0.586$\pm{}$0.042&0.589$\pm{}$0.042&0.541$\pm{}$0.044&0.581$\pm{}$0.042&0.579$\pm{}$0.043

\\ \cline{3-12}   &           & \varLL{}    &
0.542$\pm{}$0.044&0.566$\pm{}$0.043&0.580$\pm{}$0.043&0.522$\pm{}$0.050&0.539$\pm{}$0.044&0.542$\pm{}$0.048&0.491$\pm{}$0.048&0.572$\pm{}$0.044&0.566$\pm{}$0.044

\\ \cline{2-12}  & Coalition & \exLL{}     & 
0.681$\pm{}$0.035&\textBF{0.720$\pm{}$0.031}&0.685$\pm{}$0.034&0.650$\pm{}$0.037&0.655$\pm{}$0.037&\textBF{0.657$\pm{}$0.037}&\textBF{0.646$\pm{}$0.035}&0.656$\pm{}$0.037&\textBF{0.653$\pm{}$0.036}

\\ \cline{3-12}  &  & \varLL{}   &
\textBF{0.730$\pm{}$0.042}&0.546$\pm{}$0.039&0.493$\pm{}$0.047&\textBF{0.712$\pm{}$0.038}&\textBF{*0.799$\pm{}$0.024}&0.542$\pm{}$0.044&0.531$\pm{}$0.046&0.627$\pm{}$0.048&0.577$\pm{}$0.039

    \\ \hline
\end{tabular}
}
\end{table}
\begin{table}[H]
\centering
\caption{Mean and standard error of $SEQI$ (0 = worst score, 1 = perfect score) for various explanation methods with varied number of convolution layers and model capacities evaluated on (a) PRONOSTIA and (b) ZEMA datasets. Highest mean value of each column is bolded. * indicates overall highest value. }
\scriptsize  (a) PRONOSTIA
\label{table:pronostia-seqi}
\resizebox{\textwidth}{!}{
\begin{tabular}{llllllllllll}
\hline
&  &  & \multicolumn{9}{c}{Sensor Explanation Quality Index ($SEQI$)} \\ \cline{4-12} 
&  &  & \multicolumn{3}{c}{1 Convolution Layer} & \multicolumn{3}{c}{2 Convolution Layers} & \multicolumn{3}{c}{3 Convolution Layers}  \\ \cline{4-12} 

    


\multicolumn{3}{c}{Method \textbackslash{ Capacity }} & \multicolumn{1}{c}{x$\frac{1}{2}$} & \multicolumn{1}{c}{x$1$} &
\multicolumn{1}{c}{x$2$} & \multicolumn{1}{c}{x$\frac{1}{2}$} & \multicolumn{1}{c}{x$1$} & \multicolumn{1}{c}{x$2$} & \multicolumn{1}{c}{x$\frac{1}{2}$} & \multicolumn{1}{c}{x$1$} & \multicolumn{1}{c}{x$2$} \\ \hline


AE  & Central   & \exLL{} &    
0.687$\pm{}$0.031&0.696$\pm{}$0.032&0.759$\pm{}$0.025&0.608$\pm{}$0.044&0.619$\pm{}$0.043&0.616$\pm{}$0.036&0.621$\pm{}$0.041&0.631$\pm{}$0.039&0.600$\pm{}$0.043

\\ \cline{3-12}  &           & DeepLIFT & 
0.498$\pm{}$0.038&0.502$\pm{}$0.034&0.524$\pm{}$0.034&0.484$\pm{}$0.043&0.420$\pm{}$0.041&0.451$\pm{}$0.035&0.514$\pm{}$0.040&0.473$\pm{}$0.034&0.442$\pm{}$0.038

\\ \cline{3-12}  &           & GradShap & 
0.584$\pm{}$0.041&0.589$\pm{}$0.042&0.634$\pm{}$0.037&0.617$\pm{}$0.038&0.599$\pm{}$0.036&0.589$\pm{}$0.036&0.595$\pm{}$0.034&0.572$\pm{}$0.037&0.608$\pm{}$0.040

\\ \cline{2-12}  & Coalition & \exLL{}     & 
\textBF{0.766$\pm{}$0.034}&0.813$\pm{}$0.025&0.846$\pm{}$0.016&0.634$\pm{}$0.041&0.703$\pm{}$0.039&\textBF{0.708$\pm{}$0.041}&0.639$\pm{}$0.038&0.643$\pm{}$0.044&0.699$\pm{}$0.031

\\ \hline BAE & Central   & \exLL{}     &
0.641$\pm{}$0.037&0.676$\pm{}$0.030&0.715$\pm{}$0.028&0.589$\pm{}$0.042&0.620$\pm{}$0.044&0.593$\pm{}$0.039&0.641$\pm{}$0.039&0.604$\pm{}$0.043&0.621$\pm{}$0.039

\\ \cline{3-12}   &           & \varLL{}    &
0.612$\pm{}$0.037&0.624$\pm{}$0.034&0.662$\pm{}$0.029&0.488$\pm{}$0.045&0.551$\pm{}$0.044&0.494$\pm{}$0.039&\textBF{0.714$\pm{}$0.040}&0.548$\pm{}$0.039&0.497$\pm{}$0.043

\\ \cline{2-12}  & Coalition & \exLL{}     & 
0.745$\pm{}$0.038&\textBF{0.816$\pm{}$0.025}&\textBF{*0.852$\pm{}$0.016}&0.646$\pm{}$0.045&0.671$\pm{}$0.043&0.702$\pm{}$0.043&0.633$\pm{}$0.040&0.685$\pm{}$0.039&0.682$\pm{}$0.035

\\ \cline{3-12}  &  & \varLL{}   &
0.645$\pm{}$0.044&0.607$\pm{}$0.043&0.646$\pm{}$0.041&\textBF{0.767$\pm{}$0.027}&\textBF{0.712$\pm{}$0.035}&0.669$\pm{}$0.040&0.709$\pm{}$0.044&\textBF{0.776$\pm{}$0.035}&\textBF{0.741$\pm{}$0.033}

    \\ \hline
\end{tabular}
}
\end{table}
\begin{table}[H]
\centering
\scriptsize  (b) ZEMA
\resizebox{\textwidth}{!}{
\begin{tabular}{llllllllllll}
\hline
&  &  & \multicolumn{9}{c}{Sensor Explanation Quality Index ($SEQI$)} \\ \cline{4-12} 
&  &  & \multicolumn{3}{c}{1 Convolution Layer} & \multicolumn{3}{c}{2 Convolution Layers} & \multicolumn{3}{c}{3 Convolution Layers}  \\ \cline{4-12} 

    


\multicolumn{3}{c}{Method \textbackslash{ Capacity }} & \multicolumn{1}{c}{x$\frac{1}{2}$} & \multicolumn{1}{c}{x$1$} &
\multicolumn{1}{c}{x$2$} & \multicolumn{1}{c}{x$\frac{1}{2}$} & \multicolumn{1}{c}{x$1$} & \multicolumn{1}{c}{x$2$} & \multicolumn{1}{c}{x$\frac{1}{2}$} & \multicolumn{1}{c}{x$1$} & \multicolumn{1}{c}{x$2$} \\ \hline


AE  & Central   & \exLL{} &    
0.616$\pm{}$0.027&0.605$\pm{}$0.027&0.609$\pm{}$0.025&0.597$\pm{}$0.028&0.609$\pm{}$0.026&0.580$\pm{}$0.028&0.615$\pm{}$0.028&0.572$\pm{}$0.030&0.574$\pm{}$0.030

\\ \cline{3-12}  &           & DeepLIFT & 
0.463$\pm{}$0.028&0.429$\pm{}$0.029&0.417$\pm{}$0.030&0.477$\pm{}$0.026&0.472$\pm{}$0.030&0.411$\pm{}$0.031&0.464$\pm{}$0.025&0.455$\pm{}$0.029&0.467$\pm{}$0.030

\\ \cline{3-12}  &           & GradShap & 
0.321$\pm{}$0.025&0.300$\pm{}$0.023&0.306$\pm{}$0.023&0.357$\pm{}$0.026&0.373$\pm{}$0.024&0.329$\pm{}$0.024&0.398$\pm{}$0.031&0.348$\pm{}$0.026&0.385$\pm{}$0.029

\\ \cline{2-12}  & Coalition & \exLL{}     & 
0.663$\pm{}$0.026&0.689$\pm{}$0.025&\textBF{0.680$\pm{}$0.025}&0.653$\pm{}$0.026&0.663$\pm{}$0.026&0.647$\pm{}$0.027&\textBF{0.644$\pm{}$0.027}&\textBF{0.659$\pm{}$0.025}&0.625$\pm{}$0.029

\\ \hline BAE & Central   & \exLL{}     &
0.614$\pm{}$0.027&0.608$\pm{}$0.026&0.601$\pm{}$0.026&0.606$\pm{}$0.027&0.592$\pm{}$0.027&0.568$\pm{}$0.028&0.593$\pm{}$0.029&0.586$\pm{}$0.029&0.590$\pm{}$0.029

\\ \cline{3-12}   &           & \varLL{}    &
0.551$\pm{}$0.025&0.532$\pm{}$0.024&0.527$\pm{}$0.024&0.528$\pm{}$0.029&0.522$\pm{}$0.025&0.517$\pm{}$0.030&0.529$\pm{}$0.028&0.544$\pm{}$0.025&0.547$\pm{}$0.026

\\ \cline{2-12}  & Coalition & \exLL{}     & 
0.670$\pm{}$0.026&\textBF{0.698$\pm{}$0.025}&0.680$\pm{}$0.026&0.652$\pm{}$0.026&0.652$\pm{}$0.027&\textBF{0.652$\pm{}$0.026}&0.644$\pm{}$0.025&0.652$\pm{}$0.026&\textBF{0.643$\pm{}$0.026}

\\ \cline{3-12}  &  & \varLL{}   &
\textBF{0.705$\pm{}$0.025}&0.583$\pm{}$0.026&0.572$\pm{}$0.029&\textBF{0.678$\pm{}$0.025}&\textBF{*0.729$\pm{}$0.019}&0.581$\pm{}$0.027&0.563$\pm{}$0.029&0.629$\pm{}$0.029&0.604$\pm{}$0.026

    \\ \hline
\end{tabular}
}
\end{table}

\textbf{Coalitional versus centralised configuration.} Both Coalitional BAE and AE with \ExNLLShort methods are consistently better than their centralised counterparts based on $G_{SDC}$ , $G_{SSER}$ and $SEQI$ metrics as evident by the significant differences shown in \cref{fig:main-cd-diagrams}. This is further evidenced by the overall best score achievements by the Coalitional BAE and AE on both PRONOSTIA and ZEMA datasets for all metrics in  \cref{table:pronostia-gsdc}, \cref{table:pronostia-sser} and \cref{table:pronostia-seqi}. Clearly, our proposed solution improved the explanation quality by reducing correlation and enforcing independence in the explanations.

Another implication of the Coalitional BAE is its highly parallelisable computations. As smart sensor technology advances, every sensor can be installed with a BAE software agent and cooperate with other agents in the heterogeneous sensor network. This further facilitates the incorporation of BAE into a flexible multi-agent system for industrial applications where new sensors can be introduced or replaced any time \cite{yong2019multi}.

\textbf{\ExNLLBoldShort versus \VarNLLBoldShort as explanation methods.} We do not see a consistent difference between \ExNLLShort and the \VarNLLShort methods. Although \cref{fig:main-cd-diagrams} indicate that both Centralised and Coalitional BAEs with \ExNLLShort yield significant advantages over their counterparts with \VarNLLShort, when we compare only the best performances in \cref{table:pronostia-gsdc}, \cref{table:pronostia-sser} and \cref{table:pronostia-seqi}, we note that while the \ExNLLShort is better on PRONOSTIA, the  \VarNLLShort method yield better scores on ZEMA with the Coalitional BAE.

One reason for this inconsistency could be that the explanation quality depends on the dataset complexity. For instance, we observe the highest $G_{SDC}$, $G_{SSER}$ and $SEQI$ achieved on ZEMA are much lower than those of PRONOSTIA. Since there are more sensors in the ZEMA testbed (11 sensors), it poses a more difficult problem to obtain higher quality explanations as compared to PRONOSTIA testbed (2 sensors). Nonetheless, this presents a new avenue for future research by combining multiple explanations methods and also further understanding the conditions when one explanation method is better than the other. 

\textbf{BAE versus AE.} We posit that using the BAE is only clearly advantageous over the AE when we use it with the coalitional configuration. In particular, although results in \cref{fig:main-cd-diagrams} indicate that switching from a Centralised BAE to a Centralised AE may actually harm the explanation quality, this pattern is reversed when we compare Coalitional BAE against the Coalitional AE with \ExNLLShort. Despite their differences being not significant, the Coalitional BAE consistently yield better ranks than Coalitional AE with \ExNLLShort on all evaluation metrics. Moreover, the \VarNLLShort method is only possible with the BAE in which it fared the best $G_{SDC}$ and $SEQI$ scores on ZEMA when used with the coalitional configuration (\cref{table:pronostia-gsdc}(b) and  \cref{table:pronostia-seqi}(c)). 

One criticism of the BAE is that it increases the number of computations by $M$ size of the ensemble, as compared to the AE. While this is true, as embedded systems continue advancing with dedicated architectures for neural computing \cite{GONG2017384}, the BAE can take advantage over these new technologies given that the BAE can be highly parallelised since each model in the ensemble trains and predicts independently. Furthermore, the highly efficient randomised MAP sampling method further improves the practicality of obtaining explanations with BAE.

\textbf{Natural versus posthoc explanation methods.} As proposed in \cref{section:sensor-attr}, the BAE is naturally equipped with explanation methods with the \ExNLLShort and \VarNLLShort under the independent likelihood assumption. Hence, the use of posthoc explanation method may not be necessary. From \cref{fig:main-cd-diagrams}, DeepLift and GradientSHAP have significantly worst performances, with the exception of $G_{SDC}$ where Centralised BAE with \VarNLLShort is worse than the Centralised AE with GradientSHAP, but not with DeepLift. Although the difference could be due to the centralised versus coalitional configurations, to be able to compare this, we do not find a valid approach of applying the posthoc explanation methods on the Coalitional BAE and AE. The posthoc explanation methods are also much more computationally demanding as compared to simple sum of arrays with the \ExNLLShort and \VarNLLShort methods which deem them less practical for real-time IIoT applications. 

\section{Conclusion} \label{section-conc}

In this work, we presented a study on explainable HI with BAE which is a type of unsupervised DNN formulated from a probabilistic perspective. In short, our contributions can be summarised as follows: (1) proposal of two sensor attribution methods natural to the BAE based on decomposition of \ExNLLShort and \VarNLLShort, (2) the development of Coalitional BAE, (3) three quantitative metrics for evaluating quality of explanations and (4) comprehensive experiments which compared the quality of explanations using various BAE configurations and explanation methods. 

The primary implication of our study is that we gained better understanding on the explanation methods for BAE. We found that the BAE is naturally equipped with explanation methods through its probabilistic formulation. This means it is unnecessary to use posthoc explanation methods such as DeepLift and GradientSHAP which we found to be empirically underperforming compared to our proposed explanation methods. 

With the quantitative evaluation metrics that this paper has established for explainable deep learning, future studies can facilitate more rigorous comparisons between explanation methods. Specifically, the proposed metrics are (1) $G_{SDC}$ to capture the monotonic relationship between sensor attribution scores and equipment degradation, (2) $G_{SSER}$ to evaluate the ability to rank sensor importance and (3) $SEQI$ as a comprehensive measure of HI explanation quality by encompassing both $G_{SDC}$ and $G_{SSER}$ through a weighted sum. 

Through our experiments, we found that the conventional training scheme of the Centralised BAE yielded misleading explanations due to correlated outputs. For instance, we observed that non-shifting sensors were assigned monotonically increasing attribution scores and had similar shape as those of shifting sensors. This is surprising, and we caution users against being overreliant on the explanations arising from the Centralised BAEs without understanding when sensors may be failing.  

Motivated by this observation, we presented a configuration inspired by coalitional agent-based system theory called Coalitional BAE to improve the explanations of BAE's predictions. The Coalitional BAE entails training an independent BAE agent for every sensor mounted on the IIoT equipment, as opposed to training a single BAE agent for all sensor data streams. The advantage of Coalitional BAE is evident through the experiment results as we consistently find the Coalitional BAE to yield the best scores on PRONOSTIA and ZEMA datasets. 

\section*{Acknowledgment}

The work reported here was supported by European Metrology Programme for Innovation and Research (EMPIR) under the project Metrology for the Factory of the Future (MET4FOF), project number 17IND12 and part-sponsored by Research England’s Connecting Capability Fund award CCF18-7157 : Promoting the Internet of Things via Collaboration between HEIs and Industry (Pitch-In). We express our gratitude to Sascha Eichstädt for his inputs.

\bibliography{mybib, xiang}

\appendix

\section{Data pipeline using agentMET4FOF}
\label{appendix:agents}

Implementation of the data pipeline using \textit{agentMET4FOF} package is visualised and the agent roles are described. 

\begin{figure}[H]
    \begin{center}
    (a) Centralised BAE
    \centerline{\includegraphics[width=0.95\columnwidth]{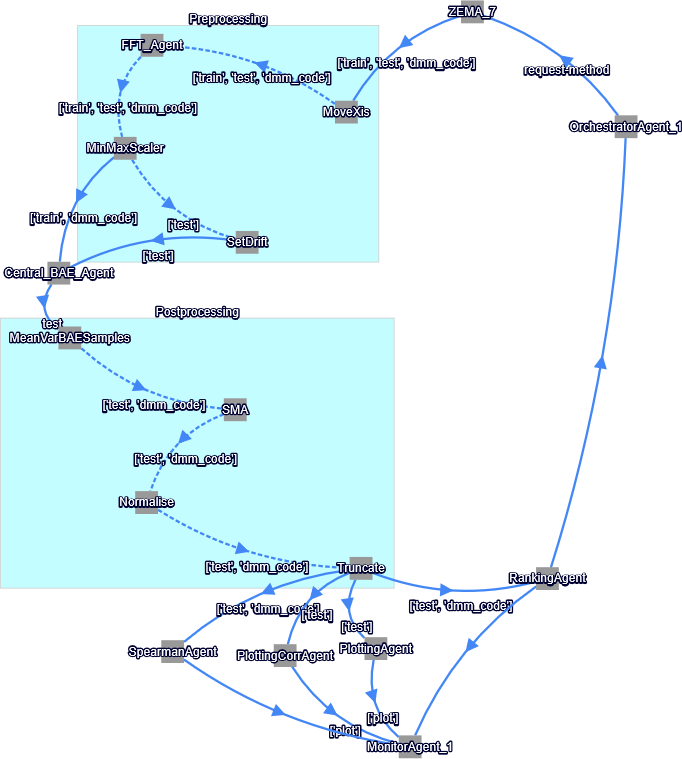}}
    \end{center}
\end{figure}

\begin{figure}[H]
    \begin{center}
    (b) Coalitional BAE
    \centerline{\includegraphics[width=\linewidth]{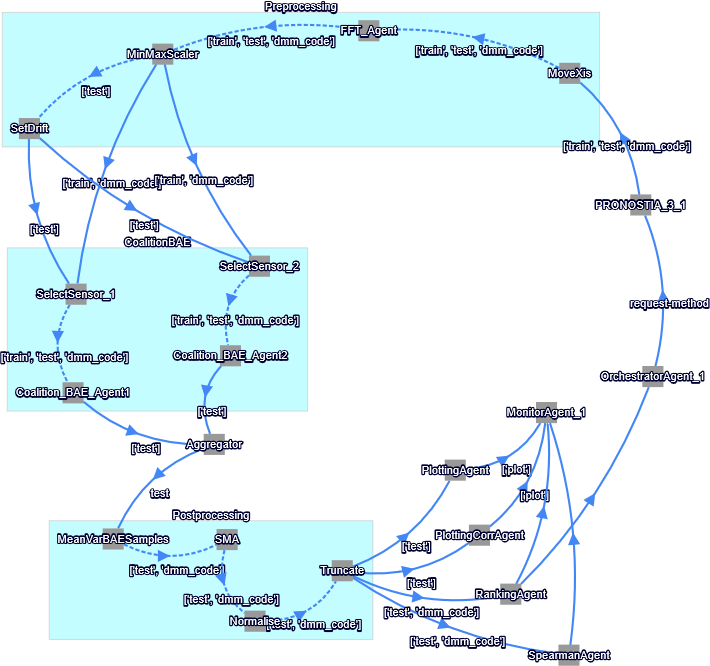}}

    \caption{\textit{agentMET4FOF} network setup for (a) Centralised BAE and (b) Coalitional BAE. Each node depicts an agent and the edges denote the channels.}
    \end{center}
\end{figure}




\begin{longtable}{p{0.2\linewidth} | p{0.75\linewidth}}
\caption{Agent classes, description of their roles}
\\
\hline
Agents & Roles \\
\hline
Datastream (e.g ZEMA\_7)                & Loads one of the run-to-failure repetition from ZEMA and PRONOSTIA data sets in an array. The first 15\% of the  trajectory data set and last 5\% were removed and splits into training and test sets.   The test set contains the full data set in which the training portion will be removed by the Truncate Agent later for a fair evaluation. \\
Move   Axis                & Swaps the array's axes. Required   to reshape the arrays from ( Cycles x Measurements x Sensors) to ( Cycles x   Sensors x Measurements).                                                                                                                                                                  \\
FFT                        & Applies fast Fourier transform   on the Measurements dimension when given an array of (Cycles x Sensors x   Measurements).                                                                                                                                                                                 \\
MinMax Scaler               & Fits and transform the Min-Max   scaler on the training data and also transforms the test data.                                                                                                                                                                                                            \\
SetDrift                   & With the specified sets of   shifting sensors and non-shifting sensors, this agent resamples the data of   the non-shifting sensors from the training set. While the shifting sensors'   data are left unchanged. This agent receives only the test set.                                                   \\
SelectSensor               & Used only for Coalitional BAE   configuration. Selects one of the sensor indexed from the (Cycles x Sensors x   Measurements) array such that an array of (Cycles x 1 x Measurements) is sent   out.                                                                                                       \\
BAE                        & Fits the BAE on the training   data. Outputs the negative log-likelihood samples in a 4D array of (BAE   Samples x Cycles x Sensors x Measurements). If posthoc explanation method is   enabled such as GradShap or DeepLift, the outputs are the attribution scores   from this method.              \\
Aggregator                 & Used only for Coalitional BAE   configuration. Aggregates and concatenates the data received from each BAE   agent such that (BAE Samples x Cycles x Sensors x Measurements) array is   recovered.                                                                                                         \\
MeanVar                    & Computes the mean and variance   of the BAE samples and subsequently sums over the Measurements dimension of   the array. Outputs the dictionary keyed by "mean" and   "var", each containing array of (Cycles x Sensors) shape.                                                                           \\
SMA       & For each Sensor, applies simple moving average to smoothen the attribution scores across the Cycles   dimension.                                                                                                                                                                                         \\
Normalise                  & Normalises the sensor   attribution scores by the mean of the training data.                                                                                                                                                                                                                               \\
Truncate                   & Truncates the training data from   the test set for evaluation.                                                                                                                                                                                                                                            \\
Plotting                   & Plots the samples as shown in   Fig. 1.                                                                                                                                                                                                                                                                  \\
PlottingCorr               & Plots the correlation between shifting and non-shifting sensors as shown in Fig. 5.                                                                                                                                                                                                                     \\
Spearman                   & Calculates the $G_{SDC}$.                                                                                                                                                                                                                                                                                  \\
Ranking                    & Calculates the $G_{SSER}$.                                                                                                                                                                                                                                                                                 \\
Monitor                    & Renders the received data on   dashboard.                                                                                                                                                                                                                                                                  \\
Orchestrator               & Omniscient agent which controls   the experiment repetitions. Varies the experimental parameters, sets of   shifting and non-shifting sensors, and the BAE configurations.                                                                                                                     \end{longtable}

\section{Additional results}
\label{appendix:additional}

Results on (1) absolute Pearson correlation coefficient between attribution scores of $S^{shift}$ and $S^{\bot{shift}}$ and (2) absolute Matthew correlation coefficient for sensor ranking (instead of $G_{SSER}$) are tabulated and compared using critical difference diagrams.

\begin{table}[H]
\centering
\caption{Mean and standard error of absolute Pearson correlation coefficient between attribution scores of $S^{shift}$ and $S^{\bot{shift}}$ for various explanation methods with varied number of convolution layers and model capacities evaluated on (a) PRONOSTIA and (b) ZEMA datasets. Highest mean value of each column is bolded. * indicates overall highest value. }

\scriptsize  (a) PRONOSTIA

\label{table:pronostia-pearson}
\resizebox{\textwidth}{!}{
\begin{tabular}{llllllllllll}
\hline
&  &  & \multicolumn{9}{c}{Pearson correlation coefficient } \\ \cline{4-12} 
&  &  & \multicolumn{3}{c}{1 Convolution Layer} & \multicolumn{3}{c}{2 Convolution Layers} & \multicolumn{3}{c}{3 Convolution Layers}  \\ \cline{4-12} 

    


\multicolumn{3}{c}{Method \textbackslash{ Capacity }} & \multicolumn{1}{c}{x$\frac{1}{2}$} & \multicolumn{1}{c}{x$1$} &
\multicolumn{1}{c}{x$2$} & \multicolumn{1}{c}{x$\frac{1}{2}$} & \multicolumn{1}{c}{x$1$} & \multicolumn{1}{c}{x$2$} & \multicolumn{1}{c}{x$\frac{1}{2}$} & \multicolumn{1}{c}{x$1$} & \multicolumn{1}{c}{x$2$} \\ \hline


AE  & Central   & \exLL{} &    
0.537$\pm{}$0.054&0.630$\pm{}$0.051&\textBF{0.634$\pm{}$0.052}&0.418$\pm{}$0.042&0.538$\pm{}$0.049&0.550$\pm{}$0.052&0.314$\pm{}$0.038&0.422$\pm{}$0.052&0.463$\pm{}$0.057

\\ \cline{3-12}  &           & DeepLIFT & 
0.501$\pm{}$0.053&0.685$\pm{}$0.054&0.644$\pm{}$0.056&0.564$\pm{}$0.057&0.615$\pm{}$0.057&0.607$\pm{}$0.059&0.372$\pm{}$0.040&0.647$\pm{}$0.054&\textBF{0.707$\pm{}$0.053}

\\ \cline{3-12}  &           & GradShap & 
0.667$\pm{}$0.035&0.605$\pm{}$0.056&0.545$\pm{}$0.063&0.575$\pm{}$0.052&0.722$\pm{}$0.041&\textBF{0.755$\pm{}$0.040}&0.573$\pm{}$0.048&0.694$\pm{}$0.047&0.691$\pm{}$0.052

\\ \cline{2-12}  & Coalition & \exLL{}     & 
0.261$\pm{}$0.037&0.242$\pm{}$0.036&0.289$\pm{}$0.032&0.298$\pm{}$0.036&0.302$\pm{}$0.030&\textBF{0.318$\pm{}$0.033}&0.290$\pm{}$0.034&0.272$\pm{}$0.032&0.312$\pm{}$0.033

\\ \hline BAE & Central   & \exLL{}     &
0.622$\pm{}$0.046&0.675$\pm{}$0.049&\textBF{0.697$\pm{}$0.046}&0.464$\pm{}$0.051&0.550$\pm{}$0.053&0.566$\pm{}$0.057&0.286$\pm{}$0.036&0.507$\pm{}$0.054&0.494$\pm{}$0.059

\\ \cline{3-12}   &           & \varLL{}    &
0.745$\pm{}$0.057&0.749$\pm{}$0.051&0.695$\pm{}$0.051&0.718$\pm{}$0.052&0.771$\pm{}$0.053&\textBF{*0.846$\pm{}$0.037}&0.388$\pm{}$0.046&0.800$\pm{}$0.045&0.773$\pm{}$0.048

\\ \cline{2-12}  & Coalition & \exLL{}     & 
0.277$\pm{}$0.036&0.295$\pm{}$0.029&0.245$\pm{}$0.030&\textBF{0.312$\pm{}$0.034}&0.287$\pm{}$0.033&0.300$\pm{}$0.034&0.288$\pm{}$0.032&0.290$\pm{}$0.032&0.305$\pm{}$0.033

\\ \cline{3-12}  &  & \varLL{}   &
0.312$\pm{}$0.037&0.268$\pm{}$0.033&0.301$\pm{}$0.031&0.304$\pm{}$0.036&0.316$\pm{}$0.034&\textBF{0.324$\pm{}$0.036}&0.270$\pm{}$0.032&0.300$\pm{}$0.032&0.298$\pm{}$0.036

    \\ \hline
\end{tabular}
}
\end{table}
\begin{table}[H]
\centering
\scriptsize  (b) ZEMA
\resizebox{\textwidth}{!}{
\begin{tabular}{llllllllllll}
\hline
&  &  & \multicolumn{9}{c}{Pearson correlation coefficient} \\ \cline{4-12} 
&  &  & \multicolumn{3}{c}{1 Convolution Layer} & \multicolumn{3}{c}{2 Convolution Layers} & \multicolumn{3}{c}{3 Convolution Layers}  \\ \cline{4-12} 
    


\multicolumn{3}{c}{Method \textbackslash{ Capacity }} & \multicolumn{1}{c}{x$\frac{1}{2}$} & \multicolumn{1}{c}{x$1$} &
\multicolumn{1}{c}{x$2$} & \multicolumn{1}{c}{x$\frac{1}{2}$} & \multicolumn{1}{c}{x$1$} & \multicolumn{1}{c}{x$2$} & \multicolumn{1}{c}{x$\frac{1}{2}$} & \multicolumn{1}{c}{x$1$} & \multicolumn{1}{c}{x$2$} \\ \hline


AE  & Central   & \exLL{} &    
0.524$\pm{}$0.035&0.515$\pm{}$0.042&0.560$\pm{}$0.043&0.517$\pm{}$0.040&0.416$\pm{}$0.038&0.469$\pm{}$0.041&0.389$\pm{}$0.033&\textBF{0.565$\pm{}$0.040}&0.395$\pm{}$0.034

\\ \cline{3-12}  &           & DeepLIFT & 
0.818$\pm{}$0.032&0.826$\pm{}$0.029&\textBF{*0.858$\pm{}$0.025}&0.788$\pm{}$0.032&0.837$\pm{}$0.028&0.823$\pm{}$0.031&0.483$\pm{}$0.041&0.778$\pm{}$0.032&0.752$\pm{}$0.039

\\ \cline{3-12}  &           & GradShap & 
0.711$\pm{}$0.032&0.699$\pm{}$0.036&0.668$\pm{}$0.037&0.678$\pm{}$0.034&0.670$\pm{}$0.033&0.700$\pm{}$0.034&0.567$\pm{}$0.035&0.676$\pm{}$0.034&\textBF{0.737$\pm{}$0.030}

\\ \cline{2-12}  & Coalition & \exLL{}     & 
0.341$\pm{}$0.026&\textBF{0.357$\pm{}$0.023}&0.325$\pm{}$0.025&0.307$\pm{}$0.028&0.341$\pm{}$0.027&0.324$\pm{}$0.029&0.328$\pm{}$0.027&0.331$\pm{}$0.026&0.339$\pm{}$0.029

\\ \hline BAE & Central   & \exLL{}     &
0.510$\pm{}$0.035&0.565$\pm{}$0.039&\textBF{0.586$\pm{}$0.040}&0.492$\pm{}$0.041&0.469$\pm{}$0.038&0.532$\pm{}$0.042&0.360$\pm{}$0.030&0.523$\pm{}$0.039&0.432$\pm{}$0.034
   
\\ \cline{3-12}   &           & \varLL{}    &
0.574$\pm{}$0.038&0.749$\pm{}$0.030&\textBF{0.816$\pm{}$0.026}&0.728$\pm{}$0.034&0.727$\pm{}$0.030&0.715$\pm{}$0.029&0.579$\pm{}$0.037&0.732$\pm{}$0.031&0.712$\pm{}$0.036

\\ \cline{2-12}  & Coalition & \exLL{}     & 
\textBF{0.338$\pm{}$0.025}&0.326$\pm{}$0.025&0.325$\pm{}$0.025&0.313$\pm{}$0.029&0.324$\pm{}$0.028&0.327$\pm{}$0.028&0.333$\pm{}$0.029&0.325$\pm{}$0.028&0.337$\pm{}$0.028

\\ \cline{3-12}  &  & \varLL{}   &
0.335$\pm{}$0.029&0.352$\pm{}$0.029&\textBF{0.373$\pm{}$0.029}&0.338$\pm{}$0.026&0.360$\pm{}$0.028&0.314$\pm{}$0.029&0.326$\pm{}$0.028&0.332$\pm{}$0.026&0.315$\pm{}$0.028

    \\ \hline
\end{tabular}
}
\end{table}

\begin{table}[H]
\centering
\caption{Mean and standard error of absolute Matthew correlation coefficient for ranking of sensors using various explanation methods with varied number of convolution layers and model capacities evaluated on (a) PRONOSTIA and (b) ZEMA datasets. Highest mean value of each column is bolded. * indicates overall highest value. }

\scriptsize  (a) PRONOSTIA

\label{table:pronostia-mcc}
\resizebox{\textwidth}{!}{
\begin{tabular}{llllllllllll}
\hline
&  &  & \multicolumn{9}{c}{Matthew correlation coefficient } \\ \cline{4-12} 
&  &  & \multicolumn{3}{c}{1 Convolution Layer} & \multicolumn{3}{c}{2 Convolution Layers} & \multicolumn{3}{c}{3 Convolution Layers}  \\ \cline{4-12} 

    


\multicolumn{3}{c}{Method \textbackslash{ Capacity }} & \multicolumn{1}{c}{x$\frac{1}{2}$} & \multicolumn{1}{c}{x$1$} &
\multicolumn{1}{c}{x$2$} & \multicolumn{1}{c}{x$\frac{1}{2}$} & \multicolumn{1}{c}{x$1$} & \multicolumn{1}{c}{x$2$} & \multicolumn{1}{c}{x$\frac{1}{2}$} & \multicolumn{1}{c}{x$1$} & \multicolumn{1}{c}{x$2$} \\ \hline


AE  & Central   & \exLL{} &    
0.754$\pm{}$0.051&0.834$\pm{}$0.046&0.926$\pm{}$0.027&0.646$\pm{}$0.053&0.694$\pm{}$0.053&0.660$\pm{}$0.059&0.633$\pm{}$0.051&0.593$\pm{}$0.052&0.614$\pm{}$0.055

\\ \cline{3-12}  &           & DeepLIFT & 
0.667$\pm{}$0.055&0.657$\pm{}$0.062&0.739$\pm{}$0.055&0.685$\pm{}$0.047&0.654$\pm{}$0.055&0.692$\pm{}$0.052&0.702$\pm{}$0.048&0.677$\pm{}$0.050&0.683$\pm{}$0.051

\\ \cline{3-12}  &           & GradShap & 
\textBF{0.934$\pm{}$0.027}&\textBF{0.898$\pm{}$0.033}&0.868$\pm{}$0.046&\textBF{0.816$\pm{}$0.048}&\textBF{0.819$\pm{}$0.052}&\textBF{0.831$\pm{}$0.044}&0.701$\pm{}$0.059&0.719$\pm{}$0.061&\textBF{0.814$\pm{}$0.050}

\\ \cline{2-12}  & Coalition & \exLL{}     & 
0.796$\pm{}$0.056&0.829$\pm{}$0.051&0.953$\pm{}$0.021&0.635$\pm{}$0.056&0.745$\pm{}$0.054&0.752$\pm{}$0.052&0.611$\pm{}$0.052&0.612$\pm{}$0.056&0.625$\pm{}$0.058

\\ \hline BAE & Central   & \exLL{}     &
0.786$\pm{}$0.048&0.821$\pm{}$0.047&0.834$\pm{}$0.050&0.656$\pm{}$0.054&0.703$\pm{}$0.052&0.668$\pm{}$0.056&0.606$\pm{}$0.053&0.639$\pm{}$0.052&0.636$\pm{}$0.056

\\ \cline{3-12}   &           & \varLL{}    &
0.791$\pm{}$0.052&0.874$\pm{}$0.037&0.825$\pm{}$0.049&0.705$\pm{}$0.054&0.746$\pm{}$0.050&0.651$\pm{}$0.062&0.699$\pm{}$0.058&0.742$\pm{}$0.054&0.664$\pm{}$0.058

\\ \cline{2-12}  & Coalition & \exLL{}     & 
0.816$\pm{}$0.049&0.897$\pm{}$0.039&\textBF{*0.971$\pm{}$0.016}&0.699$\pm{}$0.054&0.700$\pm{}$0.055&0.756$\pm{}$0.051&0.634$\pm{}$0.053&0.696$\pm{}$0.053&0.658$\pm{}$0.051

\\ \cline{3-12}  &  & \varLL{}   &
0.689$\pm{}$0.052&0.697$\pm{}$0.056&0.785$\pm{}$0.058&0.717$\pm{}$0.058&0.688$\pm{}$0.056&0.747$\pm{}$0.055&\textBF{0.771$\pm{}$0.047}&\textBF{0.819$\pm{}$0.044}&0.746$\pm{}$0.054

    \\ \hline
\end{tabular}
}
\end{table}
\begin{table}[H]
\centering
\scriptsize  (b) ZEMA
\resizebox{\textwidth}{!}{
\begin{tabular}{llllllllllll}
\hline
&  &  & \multicolumn{9}{c}{Matthew correlation coefficient} \\ \cline{4-12} 
&  &  & \multicolumn{3}{c}{1 Convolution Layer} & \multicolumn{3}{c}{2 Convolution Layers} & \multicolumn{3}{c}{3 Convolution Layers}  \\ \cline{4-12} 
    


\multicolumn{3}{c}{Method \textbackslash{ Capacity }} & \multicolumn{1}{c}{x$\frac{1}{2}$} & \multicolumn{1}{c}{x$1$} &
\multicolumn{1}{c}{x$2$} & \multicolumn{1}{c}{x$\frac{1}{2}$} & \multicolumn{1}{c}{x$1$} & \multicolumn{1}{c}{x$2$} & \multicolumn{1}{c}{x$\frac{1}{2}$} & \multicolumn{1}{c}{x$1$} & \multicolumn{1}{c}{x$2$} \\ \hline


AE  & Central   & \exLL{} &    
0.473$\pm{}$0.043&0.462$\pm{}$0.045&0.482$\pm{}$0.046&0.480$\pm{}$0.044&0.459$\pm{}$0.047&0.456$\pm{}$0.045&0.481$\pm{}$0.048&0.455$\pm{}$0.046&0.428$\pm{}$0.045

\\ \cline{3-12}  &           & DeepLIFT & 
0.399$\pm{}$0.045&0.333$\pm{}$0.045&0.327$\pm{}$0.045&0.388$\pm{}$0.044&0.396$\pm{}$0.045&0.330$\pm{}$0.044&0.310$\pm{}$0.040&0.332$\pm{}$0.044&0.371$\pm{}$0.044

\\ \cline{3-12}  &           & GradShap & 
0.207$\pm{}$0.034&0.206$\pm{}$0.035&0.213$\pm{}$0.034&0.221$\pm{}$0.034&0.220$\pm{}$0.034&0.215$\pm{}$0.034&0.271$\pm{}$0.040&0.218$\pm{}$0.036&0.259$\pm{}$0.040

\\ \cline{2-12}  & Coalition & \exLL{}     & 
0.505$\pm{}$0.050&0.551$\pm{}$0.046&\textBF{0.516$\pm{}$0.050}&0.488$\pm{}$0.050&0.505$\pm{}$0.048&\textBF{0.493$\pm{}$0.049}&\textBF{0.502$\pm{}$0.049}&0.499$\pm{}$0.049&\textBF{0.487$\pm{}$0.048}

\\ \hline BAE & Central   & \exLL{}     &
0.471$\pm{}$0.043&0.483$\pm{}$0.045&0.482$\pm{}$0.046&0.452$\pm{}$0.046&0.448$\pm{}$0.045&0.456$\pm{}$0.046&0.403$\pm{}$0.050&0.446$\pm{}$0.046&0.448$\pm{}$0.046

\\ \cline{3-12}   &           & \varLL{}    &
0.422$\pm{}$0.045&0.441$\pm{}$0.043&0.454$\pm{}$0.045&0.457$\pm{}$0.048&0.422$\pm{}$0.043&0.454$\pm{}$0.047&0.401$\pm{}$0.047&0.449$\pm{}$0.048&0.440$\pm{}$0.048

\\ \cline{2-12}  & Coalition & \exLL{}     & 
0.510$\pm{}$0.050&\textBF{0.554$\pm{}$0.047}&0.512$\pm{}$0.050&0.488$\pm{}$0.050&0.488$\pm{}$0.050&0.492$\pm{}$0.049&0.463$\pm{}$0.049&0.491$\pm{}$0.050&0.485$\pm{}$0.049

\\ \cline{3-12}  &  & \varLL{}   &
\textBF{0.654$\pm{}$0.039}&0.378$\pm{}$0.046&0.384$\pm{}$0.048&\textBF{0.579$\pm{}$0.051}&\textBF{*0.660$\pm{}$0.036}&0.406$\pm{}$0.050&0.420$\pm{}$0.046&\textBF{0.553$\pm{}$0.048}&0.408$\pm{}$0.048

    \\ \hline
\end{tabular}
}
\end{table}

\begin{figure}
    \begin{subfigure}{\textwidth}
      \centering
      \includegraphics[width=\linewidth]{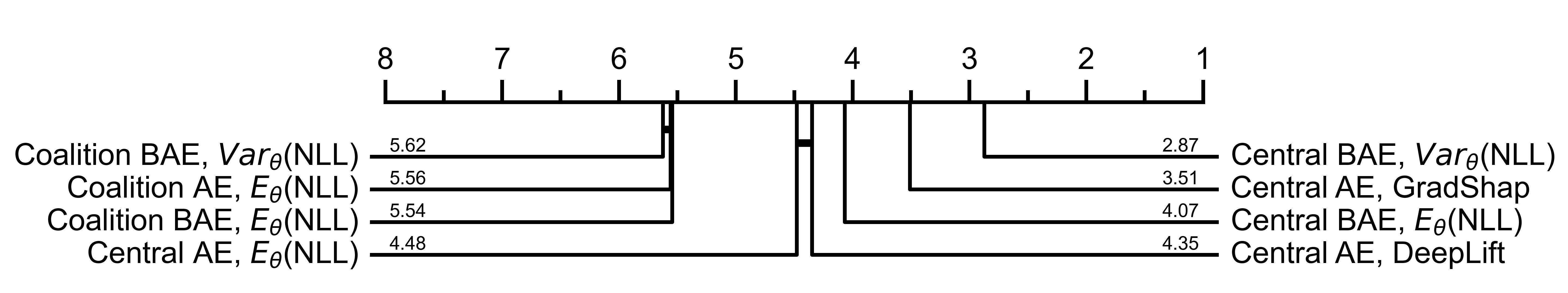}  
      \caption{Pearson correlation coefficient }
    \end{subfigure}
    \begin{subfigure}{\textwidth}
      \centering
      \includegraphics[width=\linewidth]{plots/cd-gmean-sser.png}
      \caption{Matthew correlation coefficient}
    \end{subfigure}
    \caption{Critical difference diagrams for comparing explanation methods under metrics of (a) Pearson correlation coefficient between $S^{shift}$ and $S^{\bot{shift}}$ attribution scores and (b) Matthew correlation coefficient for ranking of sensors. Average rank of each method is shown next to the labels. The methods on the right have higher ranks than the methods on left. A thick line connecting the methods indicate there is no significant difference between the methods.}
    \label{fig:appen-cd-diagrams}
\end{figure}

\end{document}